\documentclass[runningheads]{llncs}

\usepackage{eccv}

\usepackage{eccvabbrv}

\usepackage{graphicx}

\usepackage[accsupp]{axessibility}  

\usepackage[utf8]{inputenc} 
\usepackage[T1]{fontenc}    
\usepackage{url}            
\usepackage{booktabs}       
\usepackage{amsfonts}       
\usepackage{nicefrac}       
\usepackage{microtype}      
\usepackage{xcolor}         
\usepackage{color}
\usepackage{multirow}
\usepackage{wrapfig}
\usepackage{mathtools}
\usepackage{subcaption}
\usepackage{amsmath}
\usepackage{algorithm}
\usepackage{colortbl}
\usepackage{adjustbox}
\usepackage{cuted}
 \usepackage{multicol}

\definecolor{textpurple}{HTML}{883BDD} 
\RequirePackage[dvipsnames]{xcolor}

\usepackage{enumitem}

\usepackage{pifont}

\usepackage[most]{tcolorbox}

\usepackage{tcolorbox}
\usepackage{listings}
\tcbuselibrary{breakable, skins}

\newtcolorbox{promptbox}[1]{
    colback=cyan!0!white,
    colframe=cyan!60!black,
    fonttitle=\bfseries,
    title=#1,
    breakable,
    boxrule=0.8pt,
    arc=3pt,
    left=6pt, right=6pt, top=4pt, bottom=4pt
}

\lstdefinestyle{promptstyle}{
    basicstyle=\ttfamily\small,
    breaklines=true,
    backgroundcolor=\color{cyan!8!white},
    frame=none,
    xleftmargin=0pt,
    columns=flexible,
    keepspaces=true,
    literate={\\boxed}{\textbackslash boxed}6
             {<think>}{\textlangle think\textrangle}7
             {</think>}{\textlangle /think\textrangle}8
             {<answer>}{\textlangle answer\textrangle}8
             {</answer>}{\textlangle /answer\textrangle}9
}

\usepackage{hyperref}

\usepackage{orcidlink}

\begin{document}

\title{LASER: A Corrective Lens for LVLMs via Visual Attention Preservation and Sink Suppression} 
\titlerunning{LASER}

\author{Bowen Yuan\orcidlink{0009-0008-5187-4564} \and
Zijian Wang\orcidlink{0000-0002-7190-9620} \and
Yadan Luo\orcidlink{0000-0001-6272-2971} \and
Shijie Wang\orcidlink{0000-0002-7254-4715} \and
Zi Huang\orcidlink{0000-0002-9738-4949}}

\authorrunning{B.~Yuan et al.}

\institute{The University of Queensland\\
\email{\{bowen.yuan, zijian.wang, y.luo, shijie.wang, helen.huang\}@uq.edu.au}}

\maketitle

\begin{abstract}

Large vision–language models (LVLMs) exhibit strong reasoning ability but suffer from visual forgetting during long-horizon decoding, where attention progressively drifts away from visual evidence. Existing methods largely treat this issue as a late-stage attention decay problem or attempt to mitigate it through heuristic reminders or post-hoc attention lifting. Through systematic empirical analysis, we find that performance degradation under visual forgetting is largely driven by two overlooked factors: early-stage attention decay disrupts evidence acquisition, and attention concentration on a subset of task-irrelevant visual sink tokens. Motivated by these insights, we propose LASER, a post-training framework that regulates both the visual attention trajectory and intra-visual token attention distribution during reasoning. Technically, LASER introduces two complementary rewards: a Visual Grounding Reward, which encourages the model to maintain attention on semantically salient visual tokens throughout decoding, and a Sink Suppression Reward, which penalizes excessive attention concentration on visual sink tokens. Together, these rewards preserve early-stage grounding while preventing attention collapse onto uninformative regions. Extensive experiments on eight benchmark datasets demonstrate that LASER consistently outperforms strong baselines, validating attention-aware training as an effective remedy for visual forgetting. The code is available at \url{https://github.com/KeViNYuAn0314/LASER}.

\end{abstract}

\section{Introduction}
Reasoning has emerged as a transformative capability in large language models (LLMs), enabling LLMs to solve complex tasks by decomposing problems into explicit steps before producing a final answer~\cite{chain_of_though, openai_o1, least_prompt, wang2026commit, li2026can, yigeng_acl25, MATSIR_acl26}. This paradigm has been driven by reinforcement learning (RL)~\cite{deepseek_r1, openai_o1}. Rather than relying on supervised CoT data, RL incentivizes LLMs to develop sophisticated cognitive behaviors, including self-reflection, verification, and ``aha'' moments. Building on this success, recent efforts have extended RL-based reasoning to Large Vision Language Models (LVLMs), aiming to unlock analogous capabilities in the vision domain~\cite{r1_onevision, vlm-r1, vision-r1, wang2026language, fu2026mergevla}. LVLMs are trained to generate detailed chain-of-thought (CoT) rationales that jointly attend to both visual and textual information~\cite{llamav-o1, zhao2025continual, llava_cot}, yielding stronger problem-solving capabilities and improved generalization across challenging benchmarks~\cite{rl_generalize, mmeureka, du2026medfuse}.

A notable phenomenon that emerges alongside extended visual reasoning is \emph{visual forgetting}: as generation progresses, LVLMs progressively reduce attention to visual tokens and increasingly rely on self-generated textual context. Recent studies~\cite{visuothink, infi_mmr, R1_sharevl, echoes_iclr26} have linked this attention decay to performance degradation in long-horizon visual reasoning tasks. To mitigate this issue, prior work has explored two main strategies: (i) post-training methods that directly encourage sustained visual attention in the later decoding stage~\cite{reflection_v}, and (ii) introducing hand-crafted~\cite{tvc} or learned visual reminder \cite{VAPO, lookback} stages that either re-inject visual inputs (e.g., image replay or token reactivation) or prompt the model to re-focus attention on visual components during extended reasoning.

While these strategies yield empirical improvements, they primarily intervene at isolated stages without regulating attention dynamics across the reasoning trajectory. Through comprehensive empirical analysis, we argue that most of the existing works overlook two critical dimensions of preventing visual forgetting.
First, the temporal sensitivity of grounding \textit{(When)}. Our analysis reveals that the degradation often begins in early decoding stages, where visual evidence establishes the semantic scaffold for subsequent reasoning. Errors introduced at this stage would propagate autoregressively and amplify throughout the trajectory. Second, the distributional allocation of attention \textit{(Where)}. We find that even when total visual attention is preserved, attention can collapse onto a small subset of persistent sink tokens: semantically uninformative tokens that absorb disproportionate attention mass across decoding steps.
Such concentration restricts effective visual grounding. These observations suggest that mitigating visual forgetting requires both temporal preservation starting from early-stage grounding and explicit regulation of intra-visual attention allocation.

To address the limitations, we propose \textsc{LASER}, an \textbf{L}VLM \textbf{A}ttention-aware \textbf{S}ight-\textbf{E}nhanced \textbf{R}easoning framework built on Group Relative Policy Optimization (GRPO)~\cite{deepseek_r1}. \textsc{LASER} explicitly regulates both the temporal and distributional dynamics of visual attention during training. It introduces a two-component visual attention reward. The first component, a grounding preservation term, sustains early-stage non-sink visual attention and prevents premature disengagement from visual evidence. The second component, a sink suppression term, discourages disproportionate concentration on persistent sink tokens, ensuring that preserved visual attention is redistributed toward informative content. Together, these components provide a principled solution to visual forgetting in extended multimodal reasoning. The contributions of this work are threefold: 

\begin{itemize}
    \item We provide a causal analysis of visual forgetting in LVLMs, showing that performance degradation originates in early-stage grounding and that attention collapses onto persistent sink tokens, with magnitude-preserving strategies often amplifying this imbalance.
    \item We introduce \textsc{LASER}, a trajectory-level attention shaping framework that jointly regulates the temporal dynamics and intra-visual allocation of attention during long-horizon reasoning.
    \item \textsc{LASER} consistently outperforms strong baselines across benchmarks, achieving 1.9\% and 5.7\% improvements on visual reasoning and visual perception tasks, respectively. Post-hoc attention analysis shows that the improvements are accompanied by sustained visual grounding throughout generation, validating the effectiveness of our attention-centric training framework.
\end{itemize}

\begin{figure*}[!t]
\centering
    \includegraphics[width=\textwidth]{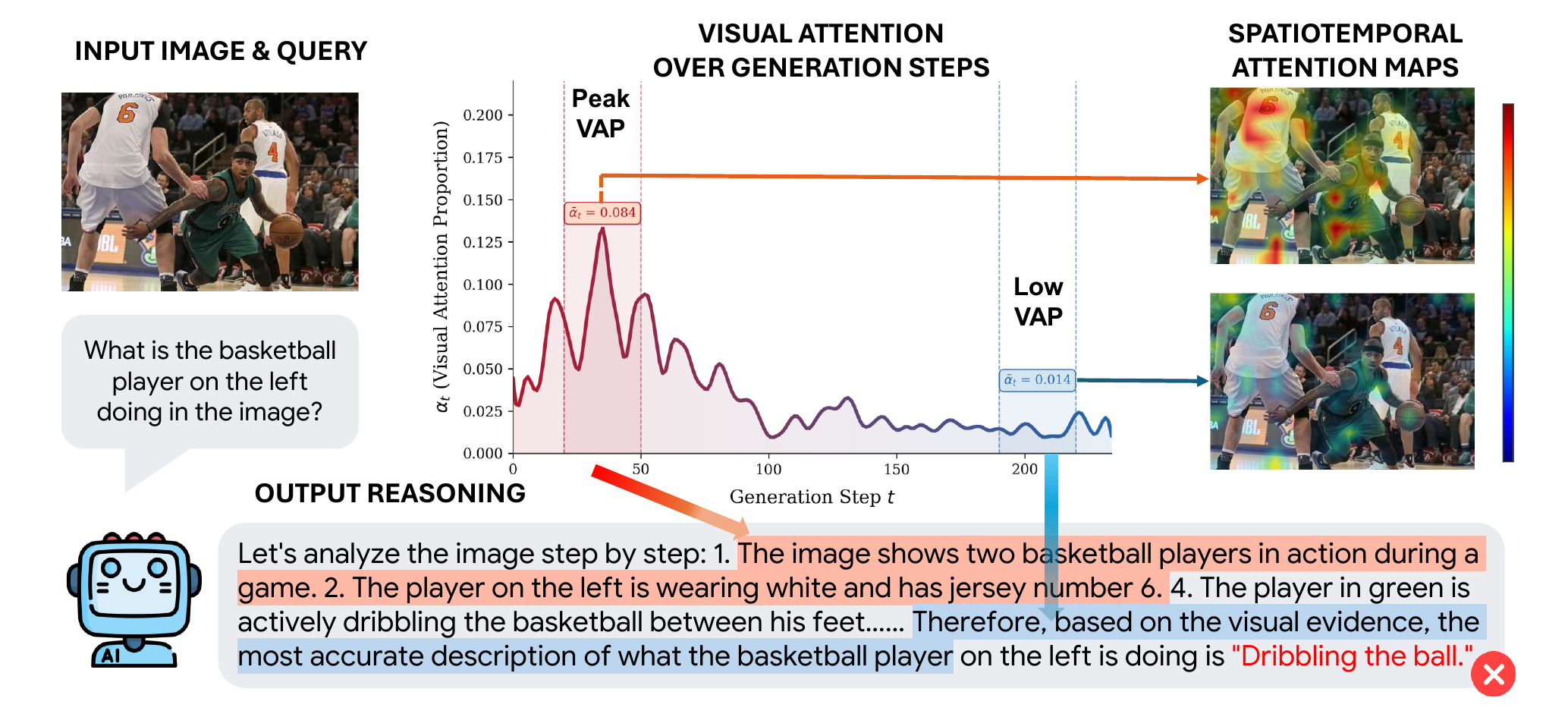}
    \caption{Illustration of visual forgetting during long-horizon reasoning in LVLMs.
\textit{Left:} an input image and the corresponding question. \textit{Middle:} the Visual Attention Proportion (VAP) over generation steps shows that attention to visual tokens peaks early and progressively decays during reasoning. \textit{Right:} difference in attention maps reveals that later decoding stages could attend less to task relevant visual regions.}
\label{fig:vis_forgetting_demo}
\end{figure*}

\section{Related Work}

\subsection{Reasoning in Large Vision-Language Models}

Large Vision-Language Models (LVLMs)~\cite{flamingo, llava, llava_next, paligemma, qwen2.5_vl} extend large language models (LLMs) with visual perception by integrating image representations into the language generation process. Given the increasing requirement for handling complex, multi-step inference tasks, eliciting LVLM reasoning capabilities has become paramount. Early works address this through chain-of-thought (CoT) prompting~\cite{DDCoT, KamCoT, VisualSketchPad} or supervised fine-tuning (SFT) on curated multimodal reasoning traces~\cite{virgo, llava_cot, llamav-o1}. While effective, SFT-based methods rely on fixed reasoning demonstrations and promote imitation, which can limit generalization to unseen reasoning patterns and complex multimodal scenarios.

To address these limitations, recent works draw inspiration from reinforcement learning (RL)–based reasoning in LLMs, notably DeepSeek-R1~\cite{deepseek_r1}, which demonstrates that Group Relative Policy Optimization (GRPO)~\cite{deepseek_math} can elicit strong reasoning behaviors through outcome-driven optimization. Building on this paradigm, several studies~\cite{r1_onevision, visionary_r1, openvlthinker, vlrethinker, vlm-r1, yuan2025wiswheat} extend RL-based training to LVLMs, enabling models to explore diverse reasoning trajectories and learn from outcome-based feedback. Specifically, Vision-SR1~\cite{Vision-SR1} introduces a self-rewarding framework that decouples visual perception from language reasoning, while VL-Rethinker~\cite{vlrethinker} addresses optimization challenges in large-scale RL via selective sample replay and forced rethinking mechanisms to promote explicit self-reflection.

\subsection{Visual Forgetting}

During extended response generation, LVLMs exhibit a notable degradation in visual grounding. Attention to visual tokens progressively diminishes, causing later responses to rely increasingly on language priors~\cite{CHAIR, fastv}. This phenomenon, referred to as \emph{visual forgetting}, has been shown to correlate with the emergence of hallucinations. Training-free techniques operate at test time by explicitly re-emphasizing visual information, such as contrastive or vision-aware decoding strategies~\cite{m3id, VCD}, or by reallocating attention mass toward visual tokens through attention manipulation~\cite{attnreal, damro}. Instruction fine-tuning approaches~\cite{MDGD, tvc} focus on preserving visual representations during text-heavy instruction tuning to reduce modality imbalance introduced by language-dominant supervision.  RL-based methods~\cite{reflection_v, VAPO} encourage visually grounded trajectories via reward design, to strengthen visual reflection during response generation.

\section{Diagnosing Visual Forgetting in LVLM Reasoning}
\label{sec:analysis}

Visual forgetting is a structural vulnerability for LVLMs during the long-horizon reasoning: as generation length increases, 
attention to visual evidence progressively 
\begin{wrapfigure}{r}{0.53\linewidth}
    \centering
    \includegraphics[width=\linewidth]{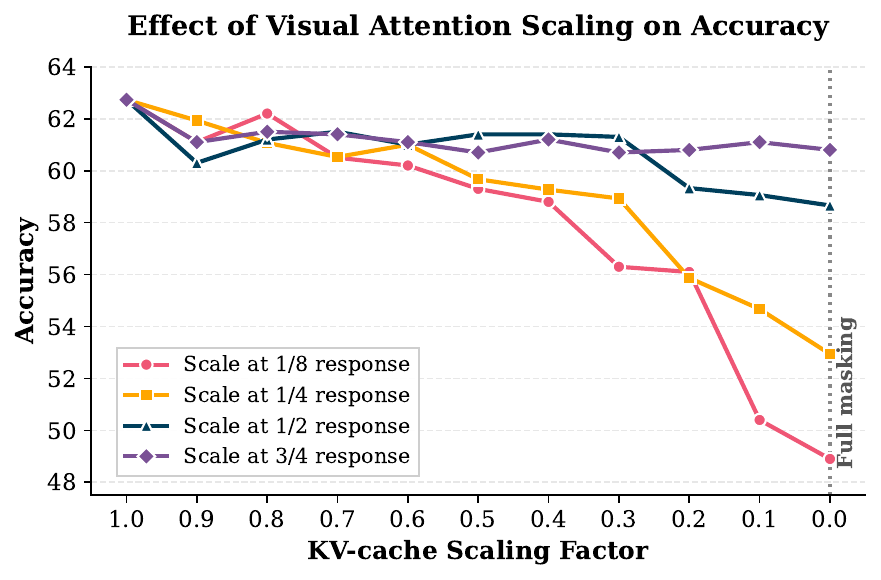}
    \caption{Accuracy on MMStar under visual KV-cache scaling factor at various reasoning stages.}
    \label{fig:v_cache_scaling_accuracy}
\end{wrapfigure}
\vspace{-1pt}
decays~\cite{tvc, reflection_v, VAPO}. However, existing characterizations primarily document the overall decay of visual attention, without examining its temporal sensitivity or internal allocation patterns. To uncover these finer-grained dynamics, we re-examine attention trajectories throughout the reasoning process. Through case studies and attention trajectory analysis (Fig.~\ref{fig:vis_forgetting_demo}), we observe that this decay is not uniform throughout the reasoning process. Early decoding steps exhibit a summarization-like role, where visual information is consolidated into intermediate textual representations. When visual grounding errors happen at this stage, errors in evidence acquisition propagate autoregressively throughout the remaining reasoning process. At the same time, we observe that the visual attention that persists is not always directed toward semantically relevant regions; a portion concentrates on visually irrelevant areas, indicating suboptimal allocation. These observations motivate a deeper investigation into two fundamental questions: \textit{when} this decay most impacts reasoning performance, and \textit{where} attention should be distributed across visual tokens to enable effective grounding.

\noindent\textbf{When is visual forgetting more consequential?}
To test the causal impact of visual attention decay, we perform controlled decoding-time interventions by applying a one-time scaling of the value states of visual tokens in the KV cache by a factor $\kappa$ at specific decoding steps. Importantly, this intervention is applied only once at the designated stage and does not affect subsequent KV updates. Because value states govern the contribution of visual tokens to downstream decoding, reducing $\kappa < 1$ transiently suppresses visual influence at that moment. We evaluate this intervention on MMStar~\cite{mmstar}, a benchmark requiring sustained visual reasoning.
As shown in Fig.~\ref{fig:v_cache_scaling_accuracy}, even a single early-stage suppression produces disproportionately large performance degradation. When full masking is applied at the $\frac{1}{8}$ stage, accuracy drops from 62.8\% to 48.9\%, whereas an identical one-time intervention at the $\frac{3}{4}$ stage results in only a 2.0\% decrease. The degradation at $\frac{1}{8}$ is over three times larger than at $\frac{1}{2}$ and nearly seven times larger than at $\frac{3}{4}$. Even under moderate suppression ($\kappa = 0.5$), early-stage intervention induces a 3.7\% absolute drop, compared to only 1.3\% at later stages. This pronounced asymmetry indicates that early-stage visual grounding plays a decisive role in forming intermediate reasoning states. Notably, despite being a one-time perturbation, early suppression induces persistent performance degradation, demonstrating that mis-grounded representations formed early continue to condition the entire trajectory.

\begin{tcolorbox}[colback=gray!10,colframe=gray!40]
\textit{Finding 1}: Visual forgetting exhibits strong early-stage sensitivity: early-stage attention decay causally impairs downstream reasoning through autoregressive propagation.
\end{tcolorbox}

\begin{figure*}[!t]
\centering
    \includegraphics[width=\textwidth]{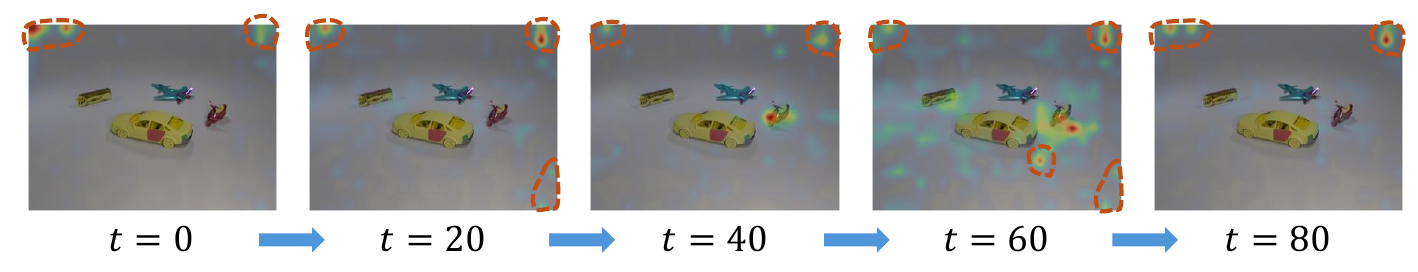}
    \caption{Visualizations of attention maps over reasoning steps. We identify image regions that receive high attention score despite lacking semantic relevance to the task.}
\label{fig:sink_example}
\end{figure*}

\begin{figure}[t]
    \centering
    \begin{subfigure}[t]{0.48\linewidth}
        \centering
        \includegraphics[width=\linewidth]{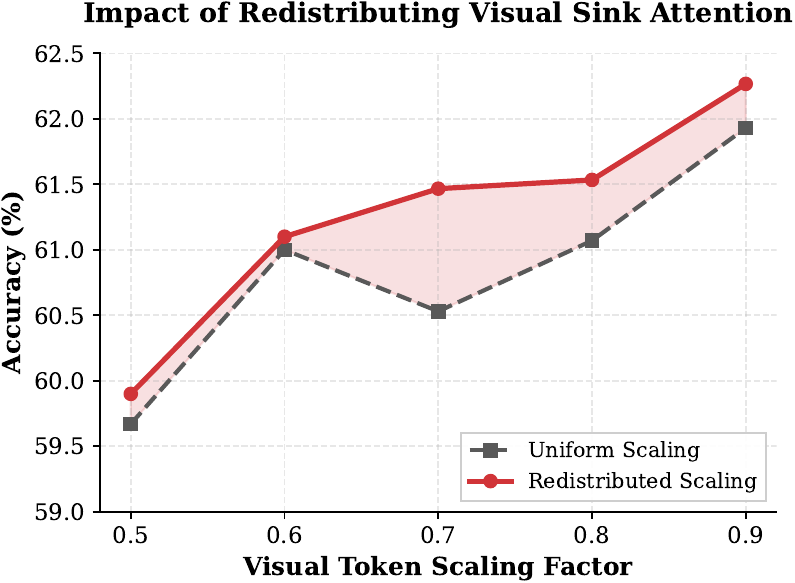}
        \label{fig:visual_sink_redist_campare}
    \end{subfigure}%
    \hfill
    \begin{subfigure}[t]{0.48\linewidth}
        \centering
        \includegraphics[width=\linewidth]{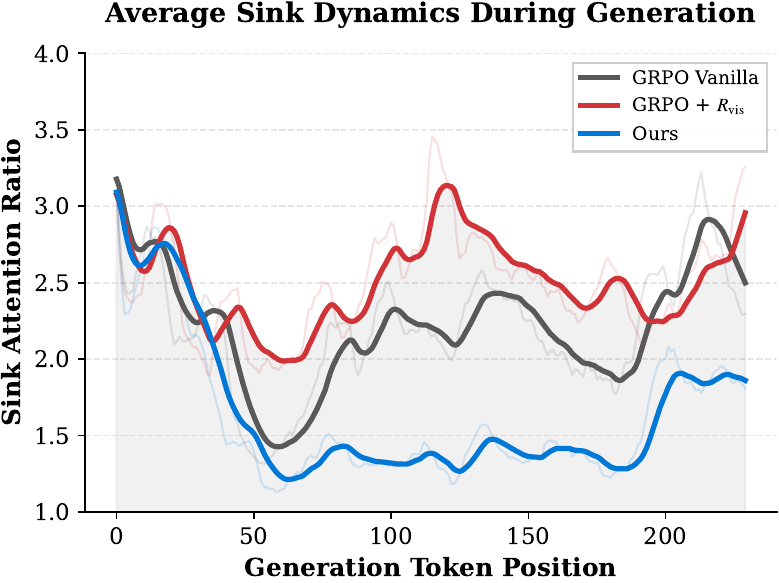}
        \label{fig:sink_attention_generation}
    \end{subfigure}
    \caption{Analysis of visual sink token attention. \textbf{Left.} Redistributing attention away from sink tokens yields consistent accuracy gains over uniform scaling. \textbf{Right.} Visual sink attention ratio across generation steps for GRPO vanilla vs. GRPO + $R_{\text{vis}}$.}
    \label{fig:sink_analysis}
\end{figure}

\label{sec:sink_finding}
\noindent\textbf{Where should visual attention be allocated?}
To investigate this question, we analyze the distribution of visual attention across individual tokens. As shown in \cref{fig:sink_example}, we observe the emergence of a persistent subset of visual tokens that consistently receive disproportionately high per-token attention across generation steps.
We refer to them as visual sink tokens $\mathcal{S} \subset \mathcal{V}$: a phenomenon~\cite{llm_stream, find_visual_sink, vtw} where certain tokens absorb disproportionate attention mass independent of semantic contribution.
Quantitatively, visual sink tokens receive $2$--$3\times$ more per-token attention than the visual-token average throughout generation (\cref{fig:sink_analysis}, right), and this concentration emerges as early as the initial decoding stages.

To assess the causal impact of distributional allocation, we compare two inference-time interventions at each scaling factor. Uniform scaling proportionally reduces the value states of all visual tokens, preserving their relative attention distribution. In contrast, redistributed scaling preserves the same total visual attention mass but reallocates attention from sink tokens to non-sink tokens.
In \cref{fig:sink_analysis} left, redistributed scaling achieves better results than uniform scaling across suppression levels. This result demonstrates that reasoning performance depends not only on the magnitude of visual attention but critically on its allocation across visual tokens. Simply preserving total attention mass is insufficient; effective grounding requires directing attention toward informative visual content.

\begin{tcolorbox}[colback=gray!10,colframe=gray!40]
\textit{Finding 2:} Visual attention is distributionally imbalanced: concentration on persistent sink tokens degrades reasoning performance even when total visual attention is preserved.
\end{tcolorbox}

Together, our findings characterize visual forgetting as a dual pathology: magnitude decay, in which global visual attention diminishes over the generation horizon, and distributional collapse, in which the remaining attention disproportionately concentrates on semantically uninformative sink tokens. Our controlled inference-time interventions demonstrate that both pathologies causally impair reasoning performance. However, such interventions operate only at inference and do not alter the model’s learned attention dynamics. As a result, they cannot provide a persistent solution that enforces reliable grounding behavior. These observations motivate a training-time approach that explicitly regulates both the temporal evolution and distributional allocation of visual attention.

\section{Methodology}
\label{sec:method}

\begin{figure*}[!t]
\centering
    \includegraphics[width=\textwidth]{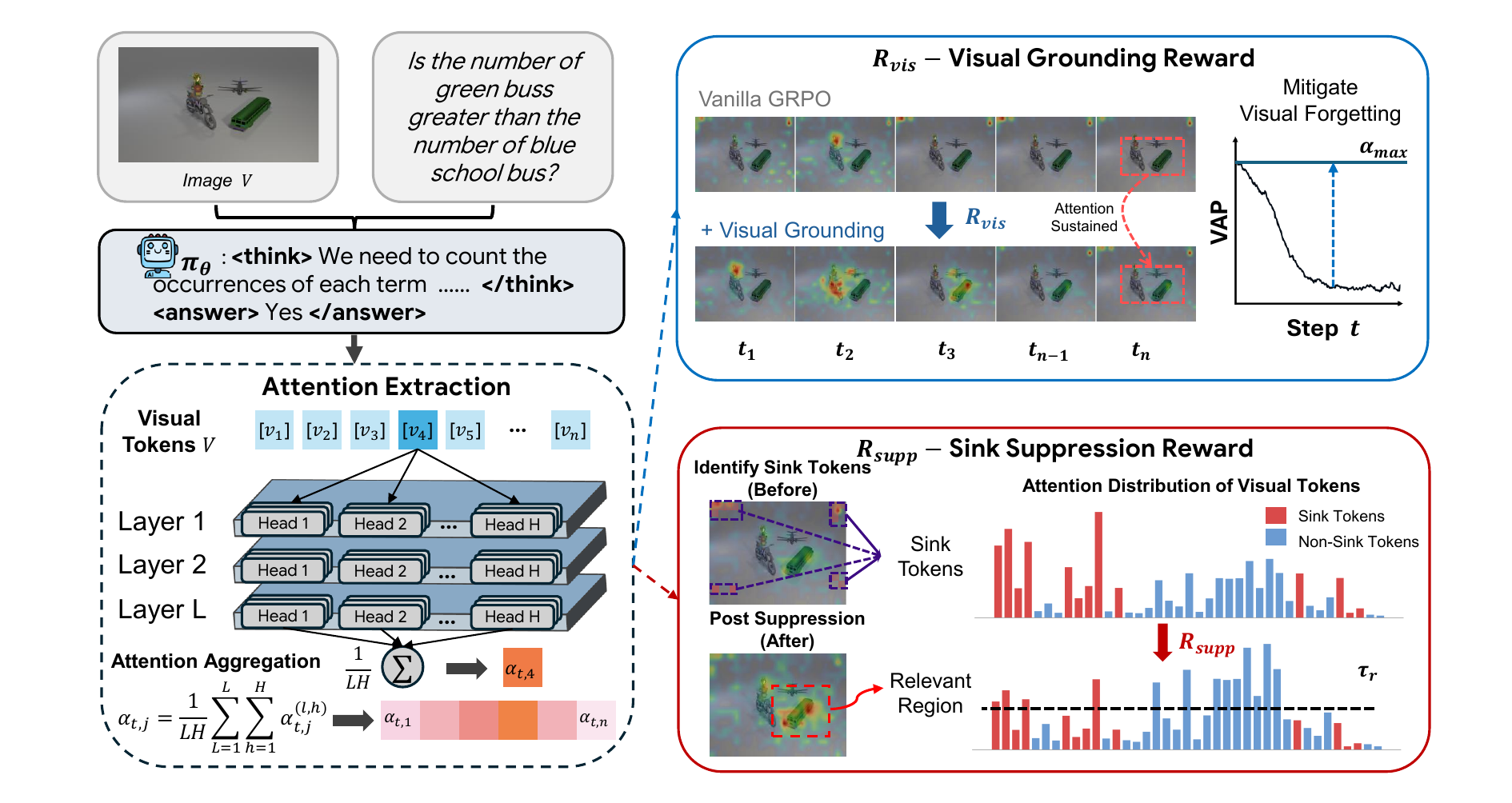}
    \caption{Overview of \textsc{LASER}. Visual token attention is extracted during generation to compute two complementary rewards: $R_\text{vis}$ sustains overall VAP across reasoning steps, while $R_{\text{supp}}$ identifies and suppresses sink tokens to rectify visual attention from visual sink tokens.}
\label{fig:framework}
\end{figure*}

Motivated by our findings of visual forgetting that stem from (i) insufficient visual attention and (ii) attention collapse to visual sink tokens, we propose \textsc{LASER}, a post-training framework that explicitly reshapes visual attention dynamics during GRPO training. Our approach introduces two complementary rewards: \emph{visual grounding reward} that sustains attention to informative visual tokens, and \emph{sink suppression reward} that penalizes excessive attention to uninformative sink tokens. The overall framework is illustrated in~\cref{fig:framework}.

\subsection{Preliminaries}
\label{sec:preliminary}

We build on Group Relative Policy Optimization (GRPO)~\cite{deepseek_math}, which eliminates the value function and estimates advantages by comparing rewards within groups of sampled outputs.
In the multimodal setting, each training instance consists of a triplet $(v, q, a)$ comprising a visual input $v$, a textual question $q$, and a ground-truth answer $a$.
The policy $\pi_{\theta}$ samples a group of $G$ independent responses $\{o_i\}_{i=1}^{G}$.
Each response $o_i$ receives a scalar reward $R_i$, and the advantage is computed via group-level normalization:
\begin{equation}
    \hat{A}_{i} = \frac{R_i - \mathrm{mean}\!\left(\{R_j\}_{j=1}^{G}\right)}{\mathrm{std}\!\left(\{R_j\}_{j=1}^{G}\right)}.
\end{equation}
GRPO updates the policy by maximizing the clipped surrogate objective with a KL penalty:
\begin{multline}
\mathcal{J}_{\mathrm{GRPO}}(\theta)
= \mathbb{E}_{\substack{(v,q,a)\sim \mathcal{D},\, \{o_i\}_{i=1}^{G} \sim \pi_{\theta_{\mathrm{old}}}(\cdot \mid v, q)}} \\
\Bigg[
\frac{1}{G}
\sum_{i=1}^{G}
\frac{1}{|o_i|}
\sum_{t=1}^{|o_i|}
\bigg(
\min\!\Big(
\rho_{i,t}(\theta)\,\hat{A}_{i},\,
\mathrm{clip}\big(\rho_{i,t}(\theta), 1-\epsilon, 1+\epsilon\big)\hat{A}_{i}
\Big)
- \beta\, D_{\mathrm{KL}}\!\left(\pi_\theta \| \pi_{\mathrm{ref}}\right)
\bigg)
\Bigg],
\end{multline}
where $\rho_{i,t}(\theta)=\pi_\theta(o_{i,t} \mid v, q, o_{i,<t})\,/\,\pi_{\theta_{\mathrm{old}}}(o_{i,t} \mid v, q, o_{i,<t})$ is the importance sampling ratio, and $\pi_{\mathrm{ref}}$ is the reference policy.

\subsection{Identifying Visual Sink Tokens}
\label{sec:sink_identification}

We establish a principled criterion to identify visual sink tokens based on the massive activation patterns observed in hidden states~\cite{sun2024massive, find_visual_sink}. Our identification process consists of two stages. First, we isolate a subset of massive activation dimensions $\mathcal{D}^{*} \subseteq [D]$, where $[D]$ denotes the full set of hidden state dimensions.
Specifically, a dimension is included in $\mathcal{D}^{*}$ if its mean absolute activation magnitude, averaged across all visual tokens and layers, anomalously exceeds the average magnitude over all dimensions.
Subsequently, a visual token $j$ is classified as a sink token if its mean activation magnitude within $\mathcal{D}^{*}$ surpasses a predefined threshold $\eta$. Formally, the set of sink tokens $\mathcal{S}$ is defined as:
\begin{equation}
\mathcal{S} = \left\{ j \in \mathcal{V} \mid \frac{1}{|\mathcal{D}^{*}|}\sum_{d \in \mathcal{D}^{*}} \bigl|h_{j,d}^{(l^*)}\bigr| > \eta \right\},
\end{equation}
where $h_{j,d}^{(l^*)}$ denotes the $d$-th component of the hidden state for token $j$ at the final layer $l^*$, and $\eta$ is determined by the top-$k$ percentile over all visual tokens $\mathcal{V}$. The remaining non-sink tokens are denoted as $\mathcal{V}' = \mathcal{V} \setminus \mathcal{S}$.

\subsection{Visual Grounding Reward}
\label{sec:grounding_reward}
To counteract the attention decay identified in \textbf{Finding 1}, we introduce a Visual Grounding Reward $R_{\text{vis}}$ that incentivizes the model to maintain focus on semantically-salient visual tokens. 
Specifically, $R_{\text{vis}}$ is formulated over the non-sink token set $\mathcal{V}'$ rather than the full set $\mathcal{V}$.
This design prevents the reinforcement of uninformative sink activations, thereby ensuring the grounding signal remains concentrated on task-critical regions and mitigating error propagation.

Mathematically, we first quantify the visual attention allocated to semantically-salient regions. At each generation step $t$, the attention proportion restricted to the non-sink token set $\mathcal{V}'$ is defined as:
\begin{equation}
   \alpha'_t = \frac{1}{LH} \sum_{l=1}^{L} \sum_{h=1}^{H} \sum_{j \in \mathcal{V}'} a_{t,j}^{(l,h)},
    \label{eq:refined_attention_ratio}
\end{equation}
where $L$ and $H$ denote the number of transformer layers and attention heads, $a_{t,j}^{(l,h)}$ is the attention weight from the generated token at step $t$ to visual token $j$ at layer $l$ and head $h$.

To mitigate the premature disengagement from visual evidence, we formulate the visual grounding reward $R_{\text{vis}}$ to penalize deviations from the peak attention level observed during the sequence. The reward is assigned as:
\begin{equation}
    R_{\text{vis}} = \sum_{t=1}^{T} \exp\!\left(-\lambda \left(1 - \frac{\alpha'_t}{\alpha'_{\max}}\right)\right) + \beta,
    \label{eq:visual_reward}
\end{equation}
where $\alpha'_{\max} = \max_t \alpha'_t$ represents the peak non-sink attention proportion, and $\lambda > 0$ governs the sensitivity to attention decay. Furthermore, we introduce $\beta$ as a penalty term to discourage reward hacking via excessively redundant reasoning sequences, thereby ensuring both grounding quality and conciseness.

\subsection{Visual Sink Suppression Reward}
\label{sec:sink_reward}


While $R_{\text{vis}}$ successfully enhances the attention weights of non-sink tokens, it remains insufficient to override the inherent attentional dominance of sink tokens. As identified in \textbf{Finding 2}, a specific subset of sink tokens exhibits excessively high activations; despite the relative gains in non-sink attention, these sink tokens continue to attract an excessive amount of the model’s focus, thereby interfering with the reasoning process. 
Hence, we introduce the Sink Suppression Reward $R_{\text{supp}}$ as a necessary constraint. Unlike $R_{\text{vis}}$, which merely promotes non-sink visual regions, $R_{\text{supp}}$ explicitly penalizes the absolute concentration on sink tokens. 


We measure this concentration as the ratio of the average per-token attention within the sink set $\mathcal{S}$ relative to the entire visual set $\mathcal{V}$:
\begin{equation}
    \bar{a}_{\mathcal{S}} = \frac{1}{|\mathcal{S}|} \sum_{j \in \mathcal{S}} a_j, \qquad
    \bar{a}_{\mathcal{V}} = \frac{1}{|\mathcal{V}|} \sum_{j \in \mathcal{V}} a_j,
\end{equation}
where $a_j$ denotes the attention weight of token $j$ averaged across all generation steps, layers, and heads. Specifically, $R_{\text{supp}}$ is formulated to suppress responses where this ratio exceeds a tolerance threshold $\tau$:
\begin{equation}
    R_{\text{supp}} = \exp\!\left(-\gamma \cdot \max\!\left(0,\; \frac{\bar{a}_{\mathcal{S}}}{\bar{a}_{\mathcal{V}}} - 1 \right)\right),
    \label{eq:sink_reward}
\end{equation}
where $\gamma > 0$ governs the penalty intensity. We set a constraint that sink tokens should not, on average, receive more attention than standard visual tokens. 

\subsection{Combined Reward Objective}
\label{sec:combined_reward}

$R_{\text{vis}}$ and $R_{\text{supp}}$ work in harmony: the former encourages sustained attention to semantically-salient regions, while the latter suppresses attention to uninformative sink tokens.
Visual rewards are only granted if the model's answer is correct.  This gating mechanism ensures that the model is rewarded for meaningful grounding rather than just increasing attention patterns on incorrect reasoning paths.
Incorporating the format check into the accuracy reward $R_{\text{acc}}$ as a necessary condition for correctness, the total reward is:
\begin{equation}
    R = R_{\text{acc}} + \omega \cdot \mathbf{1}_{[R_{\text{acc}} > 0]} \cdot \bigl(R_{\text{vis}} + R_{\text{supp}}\bigr),
    \label{eq:total_reward}
\end{equation}
where $\omega > 0$ is a single scalar that weights the attention-shaping signals.


\section{Experiment}

\subsection{Experimental Setup}
\noindent\textbf{Implementation Details.}
We use open-source LVLM Qwen-2.5-VL-7B-Instruct to evaluate our method, which is consistent with baseline methods. For cold-start, we train our model for 2 epochs following~\cite{revisual_r1}. The model is subsequently trained using GRPO with visual grounding reward and suppression reward for 2 epochs, based on verl training framework~\cite{verl}.
We source and filter our training data from the established~\cite{mmr1} and~\cite{revisual_r1}, resulting in 45K RL training samples.
All experiments are conducted on 4 NVIDIA H100 GPUs.
All experiments are evaluated using VLMEvalKit~\cite{duan2024vlmevalkit} under identical inference settings with temperature 0.6 and top-$p$ 0.95.
The detailed composition of training data and training hyperparameters is provided in supplementary materials.

\noindent\textbf{Benchmark Datasets.} We evaluate our model on 8 benchmarks to comprehensively assess visual reasoning capabilities across diverse tasks. These benchmarks span diverse domains including mathematical reasoning, general multimodal understanding, and visual perception:
\begin{itemize}
    \item Math Reasoning: MathVista~\cite{mathvista}, MathVision~\cite{mathvision}, MathVerse~\cite{mathverse}, WeMath~\cite{wemath}, which evaluate visual math problem-solving abilities.
    \item General Reasoning: MMMU~\cite{MMMU}, MMStar~\cite{mmstar}, LogicVista~\cite{logicvista}, which assess multimodal understanding and logical inference across multiple disciplines.
    \item Visual Perception: HallusionBench~\cite{hallusionbench}, which diagnoses language hallucination and visual illusion by testing model robustness to image manipulations through edited image pairs.
\end{itemize}

\noindent\textbf{Baseline Methods.} We compare our model to a diverse set of existing reasoning models. For proprietary models, we include GPT5~\cite{gpt5} and Gemini-2.5-Pro~\cite{gemini_2_5_pro} to represent the current state of the art in multimodal reasoning. We also include open-source LVLMs including InternVL2.5~\cite{internlm2} and Qwen2.5-VL~\cite{qwen2.5_vl}. Additionally, we benchmark against recent reasoning-enhanced models: VisionR1~\cite{vision-r1}, R1-Onevision~\cite{r1_onevision}, OpenVLThinker~\cite{openvlthinker}.
Among these, Reflection-V~\cite{reflection_v} and VAPO-Thinker~\cite{VAPO} are specifically designed to address visual forgetting in LVLMs. We designate Qwen2.5-VL-7B as our primary baseline to directly quantify the improvements introduced by our approach. Unless otherwise noted, baseline results are taken from the original publications; we reproduce results only when official numbers are unavailable.

\begin{table*}[t]
\centering
\caption{Performance comparison across various visual reasoning benchmarks. Best results are highlighted in \textcolor{Maroon}{\textbf{red bold}}, and the second-best results are highlighted in \textcolor{NavyBlue}{\underline{blue underlining}}.}
\resizebox{\linewidth}{!}{
    \renewcommand{\arraystretch}{1.2}
    \begin{tabular}{lcccccccc}
    \toprule
    \multirow{2}{*}{Model} & \multicolumn{4}{c}{Math Reasoning} & \multicolumn{3}{c}{General Reasoning} & \multicolumn{1}{c}{Visual} \\
    \cmidrule(lr){2-5} \cmidrule(lr){6-8} \cmidrule(lr){9-9}
     & MathVista & MathVerse & MathVision & WeMath & MMMU & MMStar & LogicVista & Hallu-Bench \\
    \midrule
    \multicolumn{9}{c}{\textit{Proprietary Model}} \\
    \midrule
    GPT5~\cite{gpt5} & 81.9 & 81.2 & 72.0 & 71.1 & 85.4 & 75.7 & 70.0 & 65.2 \\
    Gemini-2.5-Pro~\cite{gemini_2_5_pro} & 80.9 & 76.9 & 69.1 & 78.0 & 84.0 & 73.6 & 73.8 & 64.1 \\
    \midrule
    \multicolumn{9}{c}{\textit{Open-source Vision-Language Model}} \\
    \midrule
    InternVL2.5-8B~\cite{internlm2} & 68.2 & 34.3 & 25.6 & 38.6 & 49.0 & 60.8 & 38.3 & 49.0 \\
    Qwen2.5-VL-7B~\cite{qwen2.5_vl} & 66.7 & 40.7 & 26.4 & 33.1 & 52.7 & 54.9 & 42.6 & 53.6 \\
    VisionR1-7B~\cite{vision-r1} & \underline{\textcolor{NavyBlue}{73.5}} & 52.4 & 28.2 & 41.6 & 57.6 & 61.4 & 49.7 & 49.5 \\
    R1-Onevision-7B~\cite{r1_onevision} & 64.1 & 46.4 & 29.9 & \textcolor{Maroon}{\textbf{44.6}} & 54.3 & 54.1 & 45.6 & 52.5 \\
    OpenVLThinker-7B~\cite{openvlthinker} & 72.3 & 50.3 & 25.9 & - & - & 61.9 & - & 52.2 \\
    Reflection-V-7B~\cite{reflection_v} & 73.3 & - & \underline{\textcolor{NavyBlue}{33.9}} & - & \underline{\textcolor{NavyBlue}{61.3}} & - & - & 53.9 \\
    VAPO-Thinker-7B~\cite{VAPO} & \textcolor{Maroon}{\textbf{75.6}} & \underline{\textcolor{NavyBlue}{53.3}} & 31.9 & 43.6 & 60.2 & \underline{\textcolor{NavyBlue}{63.0}} & \underline{\textcolor{NavyBlue}{50.9}} & \underline{\textcolor{NavyBlue}{57.4}} \\
    \midrule
    \cellcolor{blue!10}{\textsc{LASER} (Ours)} & \cellcolor{blue!10}{72.9} & \cellcolor{blue!10}\textcolor{Maroon}{\textbf{55.2}} & \cellcolor{blue!10}\textcolor{Maroon}{\textbf{44.6}} & \cellcolor{blue!10}\underline{\textcolor{NavyBlue}{44.4}} & \cellcolor{blue!10}\textcolor{Maroon}{\textbf{62.1}} & \cellcolor{blue!10}\textcolor{Maroon}{\textbf{64.1}} &  \cellcolor{blue!10}\textcolor{Maroon}{\textbf{51.2}} &  \cellcolor{blue!10}\textcolor{Maroon}{\textbf{60.7}} \\
    \bottomrule
    \end{tabular}
}
\label{tab:main_performance}
\end{table*}


\subsection{Main Result}
As demonstrated in~\cref{tab:main_performance}, \textsc{LASER} achieves consistent performance improvements across vision-centric, hallucination, and visual reasoning benchmarks, validating the effectiveness of jointly regulating the temporal and distributional dynamics of visual attention.  Specifically, \textsc{LASER} shows a substantial 3.3\% absolute gain on HallusionBench and reaches a new best of 64.1 on MMStar, outperforming specialized methods targeting visual forgetting~\cite{VAPO, vision-r1}. \textsc{LASER} shows its strongest advantage in complex scenarios like MathVision, outperforming the closest competitor by a large margin of 12.7\%. 
The results show that these improvements are not merely incremental. The synergy between $R_{\text{vis}}$ and $R_{\text{supp}}$ ensures that task-relevant visual evidence remains active as a semantic scaffold during reasoning, while preventing attention from being dominated by uninformative sink tokens. By strengthening visual attention and maintaining sustained focus on visual inputs, our method is particularly effective on perceptually demanding tasks, demonstrating more grounded and reliable visual reasoning.

\begin{table}[t]
\centering
\caption{Ablation study with various VQA tasks on math reasoning, general visual reasoning, and visual perceptual tasks. 
Best results are highlighted in \textcolor{Maroon}{\textbf{red bold}}, and the second-best results are highlighted in \textcolor{NavyBlue}{\underline{blue underlining}}.}
\label{tab:ablation_stage}
\resizebox{0.75\linewidth}{!}{
\renewcommand{\arraystretch}{1.1}
\begin{tabular}{l|ccc|cccc}
\toprule
\textbf{Model} & \textbf{GRPO} & $R_{\text{vis}}$ & $R_{\text{supp}}$ & \textbf{Math} & \textbf{General} & \textbf{Visual} & \textbf{Avg.} \\
\midrule
\multirow{4}{*}{Qwen2.5-VL-7B~\cite{qwen2.5_vl}} & & & & 41.7 & 50.1 & 53.5 & 48.4 \\
\cline{2-8}
 & \checkmark & & & 44.0 & 54.7 & 54.6 & 51.1 \\
 & \checkmark & \checkmark & & \underline{\textcolor{NavyBlue}{45.7}} & \underline{\textcolor{NavyBlue}{55.1}} & \underline{\textcolor{NavyBlue}{55.8}} & \underline{\textcolor{NavyBlue}{52.2}} \\
 & \cellcolor{blue!10}\checkmark & \cellcolor{blue!10}\checkmark & \cellcolor{blue!10}\checkmark & \cellcolor{blue!10}\textcolor{Maroon}{\textbf{46.0}} & \cellcolor{blue!10}\textcolor{Maroon}{\textbf{58.6}} & \cellcolor{blue!10}\textcolor{Maroon}{\textbf{56.6}} & \cellcolor{blue!10}\textcolor{Maroon}{\textbf{53.7}} \\
\midrule
\multirow{3}{*}{Revisual-R1~\cite{revisual_r1}} & & & & 48.7 & 53.4 & 54.5 & 52.2 \\
\cline{2-8}
& \checkmark & & & \underline{\textcolor{NavyBlue}{52.1}} & \underline{\textcolor{NavyBlue}{57.7}} & \underline{\textcolor{NavyBlue}{59.7}} & \underline{\textcolor{NavyBlue}{56.5}} \\
& \cellcolor{blue!10}\checkmark & \cellcolor{blue!10}\checkmark & \cellcolor{blue!10}\checkmark & \cellcolor{blue!10}\textcolor{Maroon}{\textbf{54.3}} & \cellcolor{blue!10}\textcolor{Maroon}{\textbf{59.1}} & \cellcolor{blue!10}\textcolor{Maroon}{\textbf{60.7}} & \cellcolor{blue!10}\textcolor{Maroon}{\textbf{58.0}} \\
\bottomrule
\end{tabular}
}
\end{table}

\begin{figure*}[t]
    \centering

    \begin{subfigure}[t]{0.32\textwidth}
        \centering
        \includegraphics[width=\linewidth]{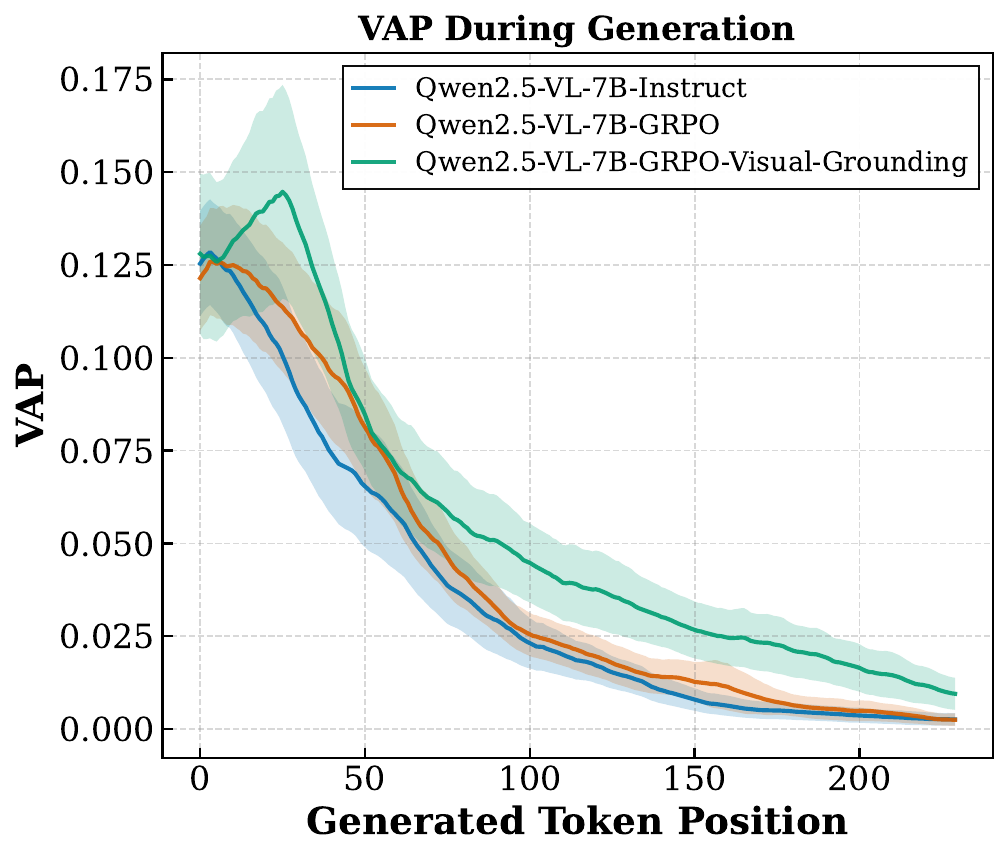}
        \caption{MathVista}
        \label{fig:attention_compare_mathvista}
    \end{subfigure}\hfill
    \begin{subfigure}[t]{0.32\textwidth}
        \centering
        \includegraphics[width=\linewidth]{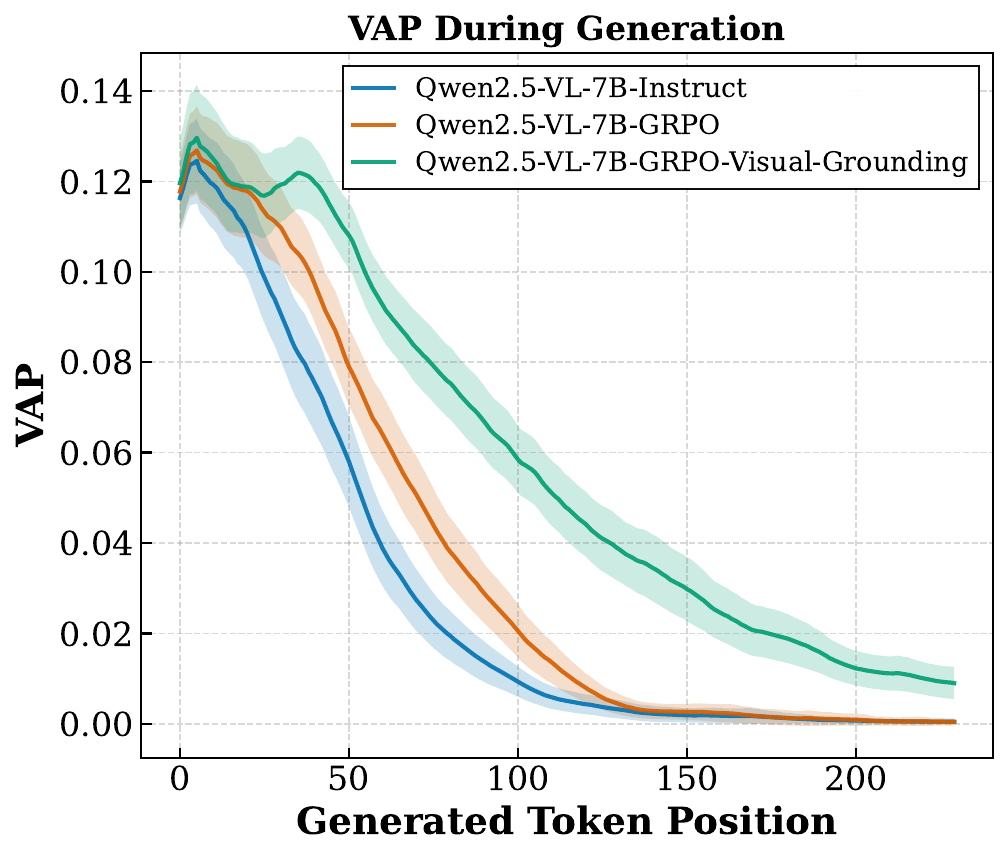}
        \caption{HallusionBench}
        \label{fig:attention_compare_hallusion}
    \end{subfigure}\hfill
    \begin{subfigure}[t]{0.32\textwidth}
        \centering
        \includegraphics[width=\linewidth]{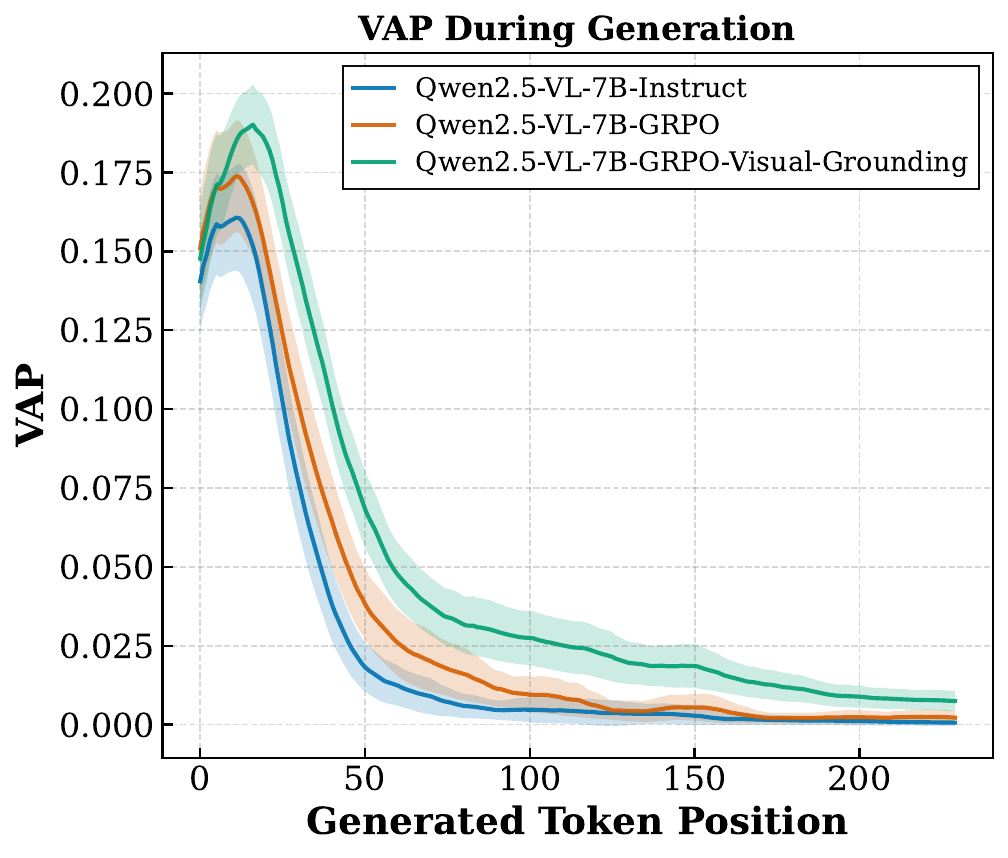}
        \caption{MMStar}
        \label{fig:attention_compare_mmstar}
    \end{subfigure}

    \caption{Analysis of VAP during generation on various datasets.}
    \label{fig:vap_three_datasets} 
\end{figure*}

\subsection{Ablation Study}

 We investigate the individual contributions of each reward within the GRPO framework by constructing incremental variants on the Qwen2.5-VL-7B base. As reported in \cref{tab:ablation_stage}, both $R_{\text{vis}}$ and $R_{\text{supp}}$ yield consistent performance gains across all benchmarks. Specifically, integrating $R_{\text{vis}}$ improves the average score from 51.1 to 52.2 by reducing visual forgetting and maintaining attention on task-relevant regions. 
 The subsequent addition of $R_{\text{supp}}$ further improves performance, particularly in general reasoning, increasing the score from 55.1 to 58.6 by reducing attention to uninformative sink tokens.
 These results demonstrate that while $R_{\text{vis}}$ ensures visual persistence throughout the reasoning trajectory, $R_{\text{supp}}$ optimizes the distributional quality of attention, collectively ensuring that visual evidence is redirected toward task-critical evidence.

To further demonstrate the effectiveness of our attention-shaping rewards, we evaluate their impact when integrated with the Supervised Fine-Tuning (SFT) base model, Revisual-R1. As observed in \cref{tab:ablation_stage}, \textsc{LASER} continues to provide substantial performance enhancements, raising the overall average from 52.2 to 58.0. 
These results show that controlling when and where the model pays visual attention improves performance on the cold-start model, demonstrating the robustness and general applicability of our approach in complex reasoning tasks.

\subsection{Further Analysis}
\noindent\textbf{\textsc{LASER} preserves visual attention throughout generation.}
\cref{fig:vap_three_datasets} presents the VAP $\alpha'_t$ across generation steps on MathVista, HallusionBench, and MMStar for three models: the Qwen2.5-VL-Instruct model, Qwen2.5-VL with GRPO and Qwen.2.5-VL trained by our method.
Across all benchmarks, the base model and vanilla GRPO exhibit lower VAP and sharp decay throughout generation, suggesting that standard RL training with outcome-based rewards provides no corrective signal for visual forgetting.
In contrast, our method consistently sustains higher VAP throughout the generation process, maintaining visual attention beyond the early generation stages where the base model and vanilla GRPO have already disengaged from visual input.


\begin{figure}[t]
    \centering
    \begin{subfigure}[t]{0.48\linewidth}
        \centering
        \includegraphics[width=\linewidth,height=4.2cm,keepaspectratio]{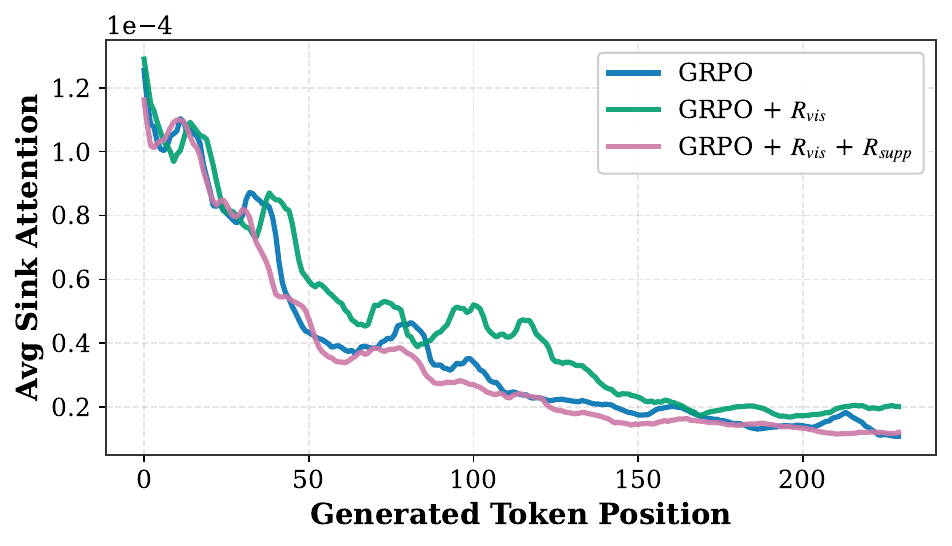}
    \end{subfigure}
    \hfill
    \begin{subfigure}[t]{0.48\linewidth}
        \centering
        \includegraphics[width=\linewidth,height=4.2cm,keepaspectratio]{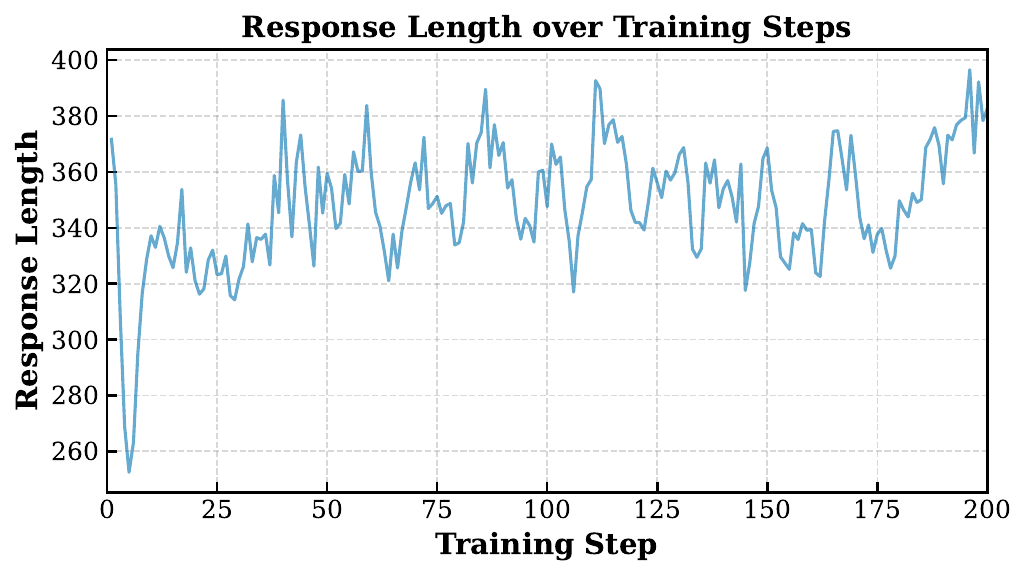}
    \end{subfigure}
    \caption{\textbf{Left.} Visual sink token attention across generation steps for model trained by \textsc{LASER}. \textbf{Right.} Response length during training process.}
    \label{fig:exp_on_sink}
\end{figure}

\noindent\textbf{\textsc{LASER} reduces disproportionate attention to visual sink tokens.}
To verify our method effectively suppresses attention allocated to visual sink tokens, we compare the average sink token attention and sink attention ratio across GRPO, GRPO with $R_{\text{vis}}$, and GRPO with both $R_{\text{vis}}$ and $R_{\text{supp}}$ (Full \textsc{LASER}).
As shown in \cref{fig:exp_on_sink} left, while $R_{\text{vis}}$ encourages the model to maintain higher overall attention on visual tokens, it simultaneously inflates attention on visual sink tokens relative to baseline. Incorporating $R_{\text{supp}}$ effectively mitigates this side effect, yielding visual sink token attention even lower than baseline throughout generation.
\cref{fig:sink_analysis} right further corroborates this through measuring the proportion of visual attention allocated to sink tokens relative to all visual tokens.
The model trained with $R_{\text{supp}}$ maintains consistently lower sink attention ratios, indicating that it redistributes attention away from semantically uninformative tokens toward more meaningful visual regions.


\noindent\textbf{Reasoning quality is maintained without length inflation or visual re-injection.}
~\cref{fig:exp_on_sink} right shows the average response length over training on Qwen2.5-VL-7B-Instruct model. Our method maintains response lengths comparable to the base model throughout training, without being prompted to generate extended chain-of-thought reasoning. This reflects that model attends more effectively to visual evidence throughout generation without incurring additional inference cost.

\begin{figure*}[!t]
\centering
    \includegraphics[width=1.02\textwidth]{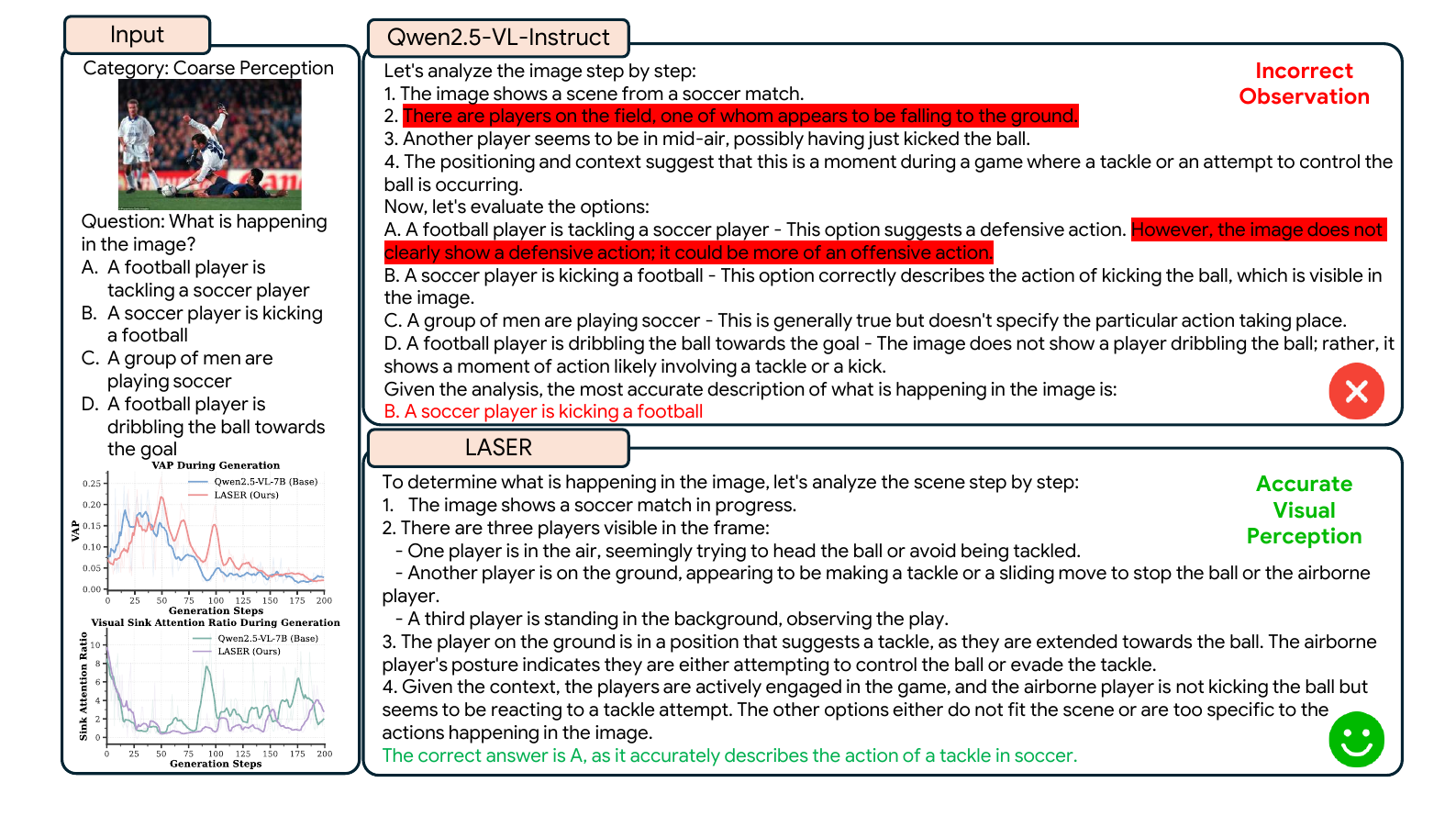}
    \caption{A qualitative comparison between Qwen2.5-VL-Instruct and \textsc{LASER} on a visual perception task. Premature visual disengagement in the base model leads to an erroneous inference, while \textsc{LASER} maintains visual attention and answers correctly.}
\label{fig:case_study}
\end{figure*}

\noindent\textbf{Sustained visual grounding improves perception accuracy.} \cref{fig:case_study} presents a representative example illustrating the improvements of our method. In this example, the base Qwen2.5-VL-Instruct model makes an erroneous summarization in its early-stage reasoning chain, resulting in an incorrect answer after long-horizon reasoning. In contrast, the model trained with \textsc{LASER} maintains fine-grained visual engagement throughout generation, and reasons coherently to the correct answer. This contrast highlights that the base model's error is not attributed to the lack of reasoning power but rather to focusing on visual evidence. By sustaining attention on semantically meaningful visual regions across generations, our method reduces the risk of hallucinated or misattributed observations.

\section{Conclusion}
In this work, we investigate the problem of visual forgetting in LVLMs, where models steadily disengage from visual evidence during extended reasoning. Through empirical analysis, we demonstrate that the onset of visual attention decay is most harmful, posing a critical impediment to reasoning performance. Beyond attention decay, we reveal a structural flaw in visual forgetting: a subset of visual sink tokens persistently dominates attention in task-irrelevant areas. Motivated by these findings, we propose \textsc{LASER}, a GRPO-based training framework that mitigates visual forgetting through two complementary rewards: a visual grounding reward that sustains attention on informative visual tokens, and a sink suppression reward that discourages spurious attention on sink tokens, steering focus toward non-sink visual evidence. Experiments show that \textsc{LASER} consistently outperforms strong baselines on visual reasoning benchmarks, effectively mitigating visual forgetting with improved visual grounding in attention.


\section*{Acknowledgements}
This research is partially supported by the Australian Research Council (IH230100013, DP230101196, DE240100105, DP240101814, IE250100108).


%
%
\bibliographystyle{splncs04}
\bibliography{main}

@String(CVPR  = {IEEE Conf. Comput. Vis. Pattern Recog.})

@String(ECCV  = {Eur. Conf. Comput. Vis.})

@String(NeurIPS = {Adv. Neural Inform. Process. Syst.})

@String(ICML  = {Int. Conf. Mach. Learn.})

@String(ICLR  = {Int. Conf. Learn. Represent.})

@String(AAAI  = {AAAI})

@String(TVC   = {The Vis. Comput.})

@String(CVPR  = {CVPR})

@String(ECCV  = {ECCV})

@String(NeurIPS = {NeurIPS})

@String(ICML  = {ICML})

@String(ICLR  = {ICLR})

@inproceedings{CLIP,
  author       = {Alec Radford and
                  Jong Wook Kim and
                  Chris Hallacy and
                  Aditya Ramesh and
                  Gabriel Goh and
                  Sandhini Agarwal and
                  Girish Sastry and
                  Amanda Askell and
                  Pamela Mishkin and
                  Jack Clark and
                  Gretchen Krueger and
                  Ilya Sutskever},
  editor       = {Marina Meila and
                  Tong Zhang},
  title        = {Learning Transferable Visual Models From Natural Language Supervision},
  booktitle    = ICML,
  pages        = {8748--8763},
  publisher    = {{PMLR}},
  year         = {2021},
}

@article{qwen2.5_vl,
  author       = {Shuai Bai and
                  Keqin Chen and
                  Xuejing Liu and
                  Jialin Wang and
                  Wenbin Ge and
                  Sibo Song and
                  Kai Dang and
                  Peng Wang and
                  Shijie Wang and
                  Jun Tang and
                  Humen Zhong and
                  Yuanzhi Zhu and
                  Ming{-}Hsuan Yang and
                  Zhaohai Li and
                  Jianqiang Wan and
                  Pengfei Wang and
                  Wei Ding and
                  Zheren Fu and
                  Yiheng Xu and
                  Jiabo Ye and
                  Xi Zhang and
                  Tianbao Xie and
                  Zesen Cheng and
                  Hang Zhang and
                  Zhibo Yang and
                  Haiyang Xu and
                  Junyang Lin},
  title        = {Qwen2.5-VL Technical Report},
  journal      = {CoRR},
  year         = {2025},
  url          = {https://doi.org/10.48550/arXiv.2502.13923},
  doi          = {10.48550/ARXIV.2502.13923},
}

@inproceedings{llava,
  author       = {Haotian Liu and
                  Chunyuan Li and
                  Qingyang Wu and
                  Yong Jae Lee},
  editor       = {Alice Oh and
                  Tristan Naumann and
                  Amir Globerson and
                  Kate Saenko and
                  Moritz Hardt and
                  Sergey Levine},
  title        = {Visual Instruction Tuning},
  booktitle    = NeurIPS,
  year         = {2023},
}

@inproceedings{llava_next,
  author       = {Haotian Liu and
                  Chunyuan Li and
                  Yuheng Li and
                  Yong Jae Lee},
  title        = {Improved Baselines with Visual Instruction Tuning},
  booktitle    = CVPR,
  publisher    = {{IEEE}},
  year         = {2024},
}

@article{paligemma,
  author       = {Lucas Beyer and
                  Andreas Steiner and
                  Andr{\'{e}} Susano Pinto and
                  Alexander Kolesnikov and
                  Xiao Wang and
                  Daniel Salz and
                  Maxim Neumann and
                  Ibrahim Alabdulmohsin and
                  Michael Tschannen and
                  Emanuele Bugliarello and
                  Thomas Unterthiner and
                  Daniel Keysers and
                  Skanda Koppula and
                  Fangyu Liu and
                  Adam Grycner and
                  Alexey A. Gritsenko and
                  Neil Houlsby and
                  Manoj Kumar and
                  Keran Rong and
                  Julian Eisenschlos and
                  Rishabh Kabra and
                  Matthias Bauer and
                  Matko Bosnjak and
                  Xi Chen and
                  Matthias Minderer and
                  Paul Voigtlaender and
                  Ioana Bica and
                  Ivana Balazevic and
                  Joan Puigcerver and
                  Pinelopi Papalampidi and
                  Olivier J. H{\'{e}}naff and
                  Xi Xiong and
                  Radu Soricut and
                  Jeremiah Harmsen and
                  Xiaohua Zhai},
  title        = {PaliGemma: {A} versatile 3B {VLM} for transfer},
  journal      = {CoRR},
  year         = {2024},
  url          = {https://doi.org/10.48550/arXiv.2407.07726},
}

@inproceedings{flamingo,
  author       = {Jean{-}Baptiste Alayrac and
                  Jeff Donahue and
                  Pauline Luc and
                  Antoine Miech and
                  Iain Barr and
                  Yana Hasson and
                  Karel Lenc and
                  Arthur Mensch and
                  Katherine Millican and
                  Malcolm Reynolds and
                  Roman Ring and
                  Eliza Rutherford and
                  Serkan Cabi and
                  Tengda Han and
                  Zhitao Gong and
                  Sina Samangooei and
                  Marianne Monteiro and
                  Jacob L. Menick and
                  Sebastian Borgeaud and
                  Andy Brock and
                  Aida Nematzadeh and
                  Sahand Sharifzadeh and
                  Mikolaj Binkowski and
                  Ricardo Barreira and
                  Oriol Vinyals and
                  Andrew Zisserman and
                  Kar{\'{e}}n Simonyan},
  title        = {Flamingo: a Visual Language Model for Few-Shot Learning},
  booktitle    = NeurIPS,
  year         = {2022},

}

@article{llava_cot,
  author       = {Guowei Xu and
                  Peng Jin and
                  Hao Li and
                  Yibing Song and
                  Lichao Sun and
                  Li Yuan},
  title        = {LLaVA-CoT: Let Vision Language Models Reason Step-by-Step},
  journal      = {CoRR},
  year         = {2024},
  url          = {https://doi.org/10.48550/arXiv.2411.10440},
}

@article{internlm2,
  author       = {Zheng Cai and
                  Maosong Cao and
                  Haojiong Chen and
                  Kai Chen and
                  Keyu Chen and
                  Xin Chen and
                  et al.},
  title        = {InternLM2 Technical Report},
  journal      = {CoRR},
  year         = {2024},
  url          = {https://doi.org/10.48550/arXiv.2403.17297},
}

@article{r1_onevision,
  author       = {Yi Yang and
                  Xiaoxuan He and
                  Hongkun Pan and
                  Xiyan Jiang and
                  Yan Deng and
                  Xingtao Yang and
                  Haoyu Lu and
                  Dacheng Yin and
                  Fengyun Rao and
                  Minfeng Zhu and
                  Bo Zhang and
                  Wei Chen},
  title        = {R1-Onevision: Advancing Generalized Multimodal Reasoning through Cross-Modal
                  Formalization},
  journal      = {CoRR},
  year         = {2025},
  url          = {https://doi.org/10.48550/arXiv.2503.10615},
}

@article{virgo,
  author       = {Yifan Du and
                  Zikang Liu and
                  Yifan Li and
                  Wayne Xin Zhao and
                  Yuqi Huo and
                  Bingning Wang and
                  Weipeng Chen and
                  Zheng Liu and
                  Zhongyuan Wang and
                  Ji{-}Rong Wen},
  title        = {Virgo: {A} Preliminary Exploration on Reproducing o1-like {MLLM}},
  journal      = {CoRR},
  year         = {2025},
  url          = {https://doi.org/10.48550/arXiv.2501.01904},
}

@article{visionary_r1,
  author       = {Jiaer Xia and
                  Yuhang Zang and
                  Peng Gao and
                  Yixuan Li and
                  Kaiyang Zhou},
  title        = {Visionary-R1: Mitigating Shortcuts in Visual Reasoning with Reinforcement
                  Learning},
  journal      = {CoRR},
  year         = {2025},
  url          = {https://doi.org/10.48550/arXiv.2505.14677},
}

@article{openvlthinker,
  author       = {Yihe Deng and
                  Hritik Bansal and
                  Fan Yin and
                  Nanyun Peng and
                  Wei Wang and
                  Kai{-}Wei Chang},
  title        = {OpenVLThinker: An Early Exploration to Complex Vision-Language Reasoning
                  via Iterative Self-Improvement},
  journal      = {CoRR},
  year         = {2025},
  url          = {https://doi.org/10.48550/arXiv.2503.17352},
}

@article{vlrethinker,
  author       = {Haozhe Wang and
                  Chao Qu and
                  Zuming Huang and
                  Wei Chu and
                  Fangzhen Lin and
                  Wenhu Chen},
  title        = {VL-Rethinker: Incentivizing Self-Reflection of Vision-Language Models
                  with Reinforcement Learning},
  journal      = {CoRR},
  year         = {2025},
  url          = {https://doi.org/10.48550/arXiv.2504.08837}
}

@article{vlm-r1,
  author       = {Haozhan Shen and
                  Peng Liu and
                  Jingcheng Li and
                  Chunxin Fang and
                  Yibo Ma and
                  Jiajia Liao and
                  Qiaoli Shen and
                  Zilun Zhang and
                  Kangjia Zhao and
                  Qianqian Zhang and
                  Ruochen Xu and
                  Tiancheng Zhao},
  title        = {{VLM-R1:} {A} Stable and Generalizable R1-style Large Vision-Language
                  Model},
  journal      = {CoRR},
  year         = {2025},
  url          = {https://doi.org/10.48550/arXiv.2504.07615},
}

@inproceedings{m3id,
  author       = {Alessandro Favero and
                  Luca Zancato and
                  Matthew Trager and
                  Siddharth Choudhary and
                  Pramuditha Perera and
                  Alessandro Achille and
                  Ashwin Swaminathan and
                  Stefano Soatto},
  title        = {Multi-Modal Hallucination Control by Visual Information Grounding},
  booktitle    = CVPR,
  pages        = {14303--14312},
  publisher    = {{IEEE}},
  year         = {2024},
}

@inproceedings{CHAIR,
  author       = {Anna Rohrbach and
                  Lisa Anne Hendricks and
                  Kaylee Burns and
                  Trevor Darrell and
                  Kate Saenko},
  editor       = {Ellen Riloff and
                  David Chiang and
                  Julia Hockenmaier and
                  Jun'ichi Tsujii},
  title        = {Object Hallucination in Image Captioning},
  booktitle    = {EMNLP},
  pages        = {4035--4045},
  publisher    = {Association for Computational Linguistics},
  year         = {2018},
}

@inproceedings{VCD,
  author       = {Sicong Leng and
                  Hang Zhang and
                  Guanzheng Chen and
                  Xin Li and
                  Shijian Lu and
                  Chunyan Miao and
                  Lidong Bing},
  title        = {Mitigating Object Hallucinations in Large Vision-Language Models through
                  Visual Contrastive Decoding},
  booktitle    = CVPR,
  pages        = {13872--13882},
  publisher    = {{IEEE}},
  year         = {2024},
}

@inproceedings{fastv,
  author       = {Liang Chen and
                  Haozhe Zhao and
                  Tianyu Liu and
                  Shuai Bai and
                  Junyang Lin and
                  Chang Zhou and
                  Baobao Chang},
  title        = {An Image is Worth 1/2 Tokens After Layer 2: Plug-and-Play Inference
                  Acceleration for Large Vision-Language Models},
  booktitle    = ECCV,
  pages        = {19--35},
  publisher    = {Springer},
  year         = {2024},
}

@article{attnreal,
  author       = {Chongjun Tu and
                  Peng Ye and
                  Dongzhan Zhou and
                  Lei Bai and
                  Gang Yu and
                  Tao Chen and
                  Wanli Ouyang},
  title        = {Attention Reallocation: Towards Zero-cost and Controllable Hallucination
                  Mitigation of MLLMs},
  journal      = {CoRR},
  year         = {2025},
  url          = {https://doi.org/10.48550/arXiv.2503.08342},
}

@inproceedings{damro,
  author       = {Xuan Gong and
                  Tianshi Ming and
                  Xinpeng Wang and
                  Zhihua Wei},
  title        = {{DAMRO:} Dive into the Attention Mechanism of {LVLM} to Reduce Object
                  Hallucination},
  booktitle    = {EMNLP},
  pages        = {7696--7712},
  publisher    = {Association for Computational Linguistics},
  year         = {2024},
}

@article{reflection_v,
  author       = {Pu Jian and
                  Junhong Wu and
                  Wei Sun and
                  Chen Wang and
                  Shuo Ren and
                  Jiajun Zhang},
  title        = {Look Again, Think Slowly: Enhancing Visual Reflection in Vision-Language
                  Models},
  journal      = {CoRR},
  year         = {2025},
  url          = {https://doi.org/10.48550/arXiv.2509.12132},
}

@article{VAPO,
  author       = {Xinyu Tian and
                  Shu Zou and
                  Zhaoyuan Yang and
                  Mengqi He and
                  Fabian Waschkowski and
                  Lukas Wesemann and
                  Peter H. Tu and
                  Jing Zhang},
  title        = {More Thought, Less Accuracy? On the Dual Nature of Reasoning in Vision-Language
                  Models},
  journal      = {CoRR},
  year         = {2025},
  url          = {https://doi.org/10.48550/arXiv.2509.25848},
}

@article{MDGD,
  author       = {Junda Wu and
                  Yuxin Xiong and
                  Xintong Li and
                  Yu Xia and
                  Ruoyu Wang and
                  Yu Wang and
                  Tong Yu and
                  Sungchul Kim and
                  Ryan A. Rossi and
                  Lina Yao and
                  Jingbo Shang and
                  Julian J. McAuley},
  title        = {Mitigating Visual Knowledge Forgetting in {MLLM} Instruction-tuning
                  via Modality-decoupled Gradient Descent},
  journal      = {CoRR},
  year         = {2025},
  url          = {https://doi.org/10.48550/arXiv.2502.11740},
}

@inproceedings{tvc,
  author       = {Hai{-}Long Sun and
                  Zhun Sun and
                  Houwen Peng and
                  Han{-}Jia Ye},
  title        = {Mitigating Visual Forgetting via Take-along Visual Conditioning for
                  Multi-modal Long CoT Reasoning},
  booktitle    = {ACL},
  pages        = {5158--5171},
  publisher    = {Association for Computational Linguistics},
  year         = {2025},
}

@misc{gpt5,
      title={OpenAI GPT-5 System Card}, 
      author={OpenAI},
      year={2025},
      url={https://arxiv.org/abs/2601.03267}, 
}

@article{gemini_2_5_pro,
  author       = {Gemini Team},
  title        = {Gemini 2.5: Pushing the Frontier with Advanced Reasoning, Multimodality,
                  Long Context, and Next Generation Agentic Capabilities},
  journal      = {CoRR},
  volume       = {abs/2507.06261},
  year         = {2025},
  url          = {https://doi.org/10.48550/arXiv.2507.06261},
}

@inproceedings{mathvista,
  author       = {Pan Lu and
                  Hritik Bansal and
                  Tony Xia and
                  Jiacheng Liu and
                  Chunyuan Li and
                  Hannaneh Hajishirzi and
                  Hao Cheng and
                  Kai{-}Wei Chang and
                  Michel Galley and
                  Jianfeng Gao},
  title        = {MathVista: Evaluating Mathematical Reasoning of Foundation Models
                  in Visual Contexts},
  booktitle    = ICLR,
  publisher    = {OpenReview.net},
  year         = {2024},
}

@inproceedings{mathvision,
  author       = {Ke Wang and
                  Junting Pan and
                  Weikang Shi and
                  Zimu Lu and
                  Houxing Ren and
                  Aojun Zhou and
                  Mingjie Zhan and
                  Hongsheng Li},
  title        = {Measuring Multimodal Mathematical Reasoning with MATH-Vision Dataset},
  booktitle    = Neurips,
  year         = {2024},
}

@inproceedings{mathverse,
  author       = {Renrui Zhang and
                  Dongzhi Jiang and
                  Yichi Zhang and
                  Haokun Lin and
                  Ziyu Guo and
                  Pengshuo Qiu and
                  Aojun Zhou and
                  Pan Lu and
                  Kai{-}Wei Chang and
                  Yu Qiao and
                  Peng Gao and
                  Hongsheng Li},
  title        = {{MATHVERSE:} Does Your Multi-modal {LLM} Truly See the Diagrams in
                  Visual Math Problems?},
  booktitle    = ECCV,
  pages        = {169--186},
  publisher    = {Springer},
  year         = {2024},
}

@inproceedings{wemath,
  author       = {Runqi Qiao and
                  Qiuna Tan and
                  Guanting Dong and
                  Minhui Wu and
                  Chong Sun and
                  Xiaoshuai Song and
                  Jiapeng Wang and
                  Zhuoma Gongque and
                  Shanglin Lei and
                  Yifan Zhang and
                  Zhe Wei and
                  Miaoxuan Zhang and
                  Runfeng Qiao and
                  Xiao Zong and
                  Yida Xu and
                  Peiqing Yang and
                  Zhimin Bao and
                  Muxi Diao and
                  Chen Li and
                  Honggang Zhang},
  title        = {We-Math: Does Your Large Multimodal Model Achieve Human-like Mathematical
                  Reasoning?},
  booktitle    = {ACL},
  pages        = {20023--20070},
  publisher    = {Association for Computational Linguistics},
  year         = {2025},
}

@inproceedings{MMMU,
  author       = {Xiang Yue and
                  Yuansheng Ni and
                  Tianyu Zheng and
                  Kai Zhang and
                  Ruoqi Liu and
                  Ge Zhang and
                  Samuel Stevens and
                  Dongfu Jiang and
                  Weiming Ren and
                  Yuxuan Sun and
                  Cong Wei and
                  Botao Yu and
                  Ruibin Yuan and
                  Renliang Sun and
                  Ming Yin and
                  Boyuan Zheng and
                  Zhenzhu Yang and
                  Yibo Liu and
                  Wenhao Huang and
                  Huan Sun and
                  Yu Su and
                  Wenhu Chen},
  title        = {{MMMU:} {A} Massive Multi-Discipline Multimodal Understanding and
                  Reasoning Benchmark for Expert {AGI}},
  booktitle    = CVPR,
  pages        = {9556--9567},
  publisher    = {{IEEE}},
  year         = {2024},
}

@inproceedings{mmstar,
  author       = {Lin Chen and
                  Jinsong Li and
                  Xiaoyi Dong and
                  Pan Zhang and
                  Yuhang Zang and
                  Zehui Chen and
                  Haodong Duan and
                  Jiaqi Wang and
                  Yu Qiao and
                  Dahua Lin and
                  Feng Zhao},
  title        = {Are We on the Right Way for Evaluating Large Vision-Language Models?},
  booktitle    = Neurips,
  year         = {2024},
}

@article{logicvista,
  author       = {Yijia Xiao and
                  Edward Sun and
                  Tianyu Liu and
                  Wei Wang},
  title        = {LogicVista: Multimodal {LLM} Logical Reasoning Benchmark in Visual
                  Contexts},
  journal      = {CoRR},
  volume       = {abs/2407.04973},
  year         = {2024},
  url          = {https://doi.org/10.48550/arXiv.2407.04973},
}

@inproceedings{hallusionbench,
  author       = {Tianrui Guan and
                  Fuxiao Liu and
                  Xiyang Wu and
                  Ruiqi Xian and
                  Zongxia Li and
                  Xiaoyu Liu and
                  Xijun Wang and
                  Lichang Chen and
                  Furong Huang and
                  Yaser Yacoob and
                  Dinesh Manocha and
                  Tianyi Zhou},
  title        = {Hallusionbench: An Advanced Diagnostic Suite for Entangled Language
                  Hallucination and Visual Illusion in Large Vision-Language Models},
  booktitle    = CVPR,
  publisher    = {{IEEE}},
  year         = {2024},
}

@article{vision-r1,
  author       = {Wenxuan Huang and
                  Bohan Jia and
                  Zijie Zhai and
                  Shaosheng Cao and
                  Zheyu Ye and
                  Fei Zhao and
                  Zhe Xu and
                  Yao Hu and
                  Shaohui Lin},
  title        = {Vision-R1: Incentivizing Reasoning Capability in Multimodal Large
                  Language Models},
  journal      = {CoRR},
  year         = {2025},
  url          = {https://doi.org/10.48550/arXiv.2503.06749},
}

@inproceedings{VisualSketchPad,
  author       = {Yushi Hu and
                  Weijia Shi and
                  Xingyu Fu and
                  Dan Roth and
                  Mari Ostendorf and
                  Luke Zettlemoyer and
                  Noah A. Smith and
                  Ranjay Krishna},
  title        = {Visual Sketchpad: Sketching as a Visual Chain of Thought for Multimodal
                  Language Models},
  booktitle    = Neurips,
  year         = {2024},
}

@inproceedings{DDCoT,
  author       = {Ge Zheng and
                  Bin Yang and
                  Jiajin Tang and
                  Hong{-}Yu Zhou and
                  Sibei Yang},
  title        = {DDCoT: Duty-Distinct Chain-of-Thought Prompting for Multimodal Reasoning
                  in Language Models},
  booktitle    = Neurips,
  year         = {2023},
}

@inproceedings{KamCoT,
  author       = {Debjyoti Mondal and
                  Suraj Modi and
                  Subhadarshi Panda and
                  Rituraj Singh and
                  Godawari Sudhakar Rao},
  title        = {KAM-CoT: Knowledge Augmented Multimodal Chain-of-Thoughts Reasoning},
  booktitle    = AAAI,
  pages        = {18798--18806},
  publisher    = {{AAAI} Press},
  year         = {2024},
}

@inproceedings{llamav-o1,
  author       = {Omkar Thawakar and
                  Dinura Dissanayake and
                  Ketan Pravin More and
                  Ritesh Thawkar and
                  Ahmed Heakl and
                  Noor Ahsan and
                  Yuhao Li and
                  Mohammed Zumri and
                  Jean Lahoud and
                  Rao Muhammad Anwer and
                  Hisham Cholakkal and
                  Ivan Laptev and
                  Mubarak Shah and
                  Fahad Shahbaz Khan and
                  Salman H. Khan},
  title        = {LlamaV-o1: Rethinking Step-by-step Visual Reasoning in LLMs},
  booktitle    = {ACL},
  pages        = {24290--24315},
  publisher    = {Association for Computational Linguistics},
  year         = {2025},
}

@article{deepseek_r1,
  author       = {DeepSeek{-}AI},
  title        = {DeepSeek-R1: Incentivizing Reasoning Capability in LLMs via Reinforcement
                  Learning},
  journal      = {CoRR},
  volume       = {abs/2501.12948},
  year         = {2025},
  url          = {https://doi.org/10.48550/arXiv.2501.12948},
}

@article{Vision-SR1,
  author       = {Zongxia Li and
                  Wenhao Yu and
                  Chengsong Huang and
                  Rui Liu and
                  Zhenwen Liang and
                  Fuxiao Liu and
                  Jingxi Che and
                  Dian Yu and
                  Jordan L. Boyd{-}Graber and
                  Haitao Mi and
                  Dong Yu},
  title        = {Self-Rewarding Vision-Language Model via Reasoning Decomposition},
  journal      = {CoRR},
  year         = {2025},
  url          = {https://doi.org/10.48550/arXiv.2508.19652},
}

@article{deepseek_math,
  author       = {Zhihong Shao and
                  Peiyi Wang and
                  Qihao Zhu and
                  Runxin Xu and
                  Junxiao Song and
                  Mingchuan Zhang and
                  Y. K. Li and
                  Y. Wu and
                  Daya Guo},
  title        = {DeepSeekMath: Pushing the Limits of Mathematical Reasoning in Open
                  Language Models},
  journal      = {CoRR},
  year         = {2024},
  url          = {https://doi.org/10.48550/arXiv.2402.03300},
}

@inproceedings{find_visual_sink,
  author       = {Seil Kang and
                  Jinyeong Kim and
                  Junhyeok Kim and
                  Seong Jae Hwang},
  title        = {See What You Are Told: Visual Attention Sink in Large Multimodal Models},
  booktitle    = ICLR,
  year         = {2025},
  }

@article{revisual_r1,
  title={Advancing multimodal reasoning: From optimized cold start to staged reinforcement learning},
  author={Chen, Shuang and Guo, Yue and Su, Zhaochen and Li, Yafu and Wu, Yulun and Chen, Jiacheng and Chen, Jiayu and Wang, Weijie and Qu, Xiaoye and Cheng, Yu},
  journal={arXiv preprint arXiv:2506.04207},
  year={2025}
}

@misc{verl,
  author = {Volcano Engine},
  title = {verl: Volcano Engine Reinforcement Learning for LLMs},
  year = {2025},
  journal = {GitHub repository},
  howpublished = {\url{https://github.com/verl-project/verl}},
}

@article{chain_of_though,
  title={Chain-of-thought prompting elicits reasoning in large language models},
  author={Wei, Jason and Wang, Xuezhi and Schuurmans, Dale and Bosma, Maarten and Xia, Fei and Chi, Ed and Le, Quoc V and Zhou, Denny and others},
  journal=NeurIPS,
  volume={35},
  pages={24824--24837},
  year={2022}
}

@article{openai_o1,
  title={Openai o1 system card},
  author={Jaech, Aaron and Kalai, Adam and Lerer, Adam and Richardson, Adam and El-Kishky, Ahmed and Low, Aiden and Helyar, Alec and Madry, Aleksander and Beutel, Alex and Carney, Alex and others},
  journal={arXiv preprint arXiv:2412.16720},
  year={2024}
}

@article{least_prompt,
  title={Least-to-most prompting enables complex reasoning in large language models},
  author={Zhou, Denny and Sch{\"a}rli, Nathanael and Hou, Le and Wei, Jason and Scales, Nathan and Wang, Xuezhi and Schuurmans, Dale and Cui, Claire and Bousquet, Olivier and Le, Quoc and others},
  journal={arXiv preprint arXiv:2205.10625},
  year={2022}
}

@article{rl_generalize,
  title={Sft memorizes, rl generalizes: A comparative study of foundation model post-training},
  author={Chu, Tianzhe and Zhai, Yuexiang and Yang, Jihan and Tong, Shengbang and Xie, Saining and Schuurmans, Dale and Le, Quoc V and Levine, Sergey and Ma, Yi},
  journal={arXiv preprint arXiv:2501.17161},
  year={2025}
}

@article{mmeureka,
  title={Mm-eureka: Exploring the frontiers of multimodal reasoning with rule-based reinforcement learning},
  author={Meng, Fanqing and Du, Lingxiao and Liu, Zongkai and Zhou, Zhixiang and Lu, Quanfeng and Fu, Daocheng and Han, Tiancheng and Shi, Botian and Wang, Wenhai and He, Junjun and others},
  journal={arXiv preprint arXiv:2503.07365},
  year={2025}
}

@article{llm_stream,
  title={Efficient streaming language models with attention sinks},
  author={Xiao, Guangxuan and Tian, Yuandong and Chen, Beidi and Han, Song and Lewis, Mike},
  journal={arXiv preprint arXiv:2309.17453},
  year={2023}
}

@article{mmr1,
  title={Mmr1: Enhancing multimodal reasoning with variance-aware sampling and open resources},
  author={Leng, Sicong and Wang, Jing and Li, Jiaxi and Zhang, Hao and Hu, Zhiqiang and Zhang, Boqiang and Jiang, Yuming and Zhang, Hang and Li, Xin and Bing, Lidong and others},
  journal={arXiv preprint arXiv:2509.21268},
  year={2025}
}

@article{lookback,
  title={Look-back: Implicit visual re-focusing in mllm reasoning},
  author={Yang, Shuo and Niu, Yuwei and Liu, Yuyang and Ye, Yang and Lin, Bin and Yuan, Li},
  journal={arXiv preprint arXiv:2507.03019},
  year={2025}
}

@article{R1_sharevl,
  title={R1-sharevl: Incentivizing reasoning capability of multimodal large language models via share-grpo},
  author={Yao, Huanjin and Yin, Qixiang and Zhang, Jingyi and Yang, Min and Wang, Yibo and Wu, Wenhao and Su, Fei and Shen, Li and Qiu, Minghui and Tao, Dacheng and others},
  journal={arXiv preprint arXiv:2505.16673},
  year={2025}
}

@article{infi_mmr,
  title={Infi-mmr: Curriculum-based unlocking multimodal reasoning via phased reinforcement learning in multimodal small language models},
  author={Liu, Zeyu and Liu, Yuhang and Zhu, Guanghao and Xie, Congkai and Li, Zhen and Yuan, Jianbo and Wang, Xinyao and Li, Qing and Cheung, Shing-Chi and Zhang, Shengyu and others},
  journal={arXiv preprint arXiv:2505.23091},
  year={2025}
}

@inproceedings{visuothink,
  title={Visuothink: Empowering lvlm reasoning with multimodal tree search},
  author={Wang, Yikun and Wang, Siyin and Cheng, Qinyuan and Fei, Zhaoye and Ding, Liang and Guo, Qipeng and Tao, Dacheng and Qiu, Xipeng},
  booktitle={Proceedings of the 63rd Annual Meeting of the Association for Computational Linguistics (Volume 1: Long Papers)},
  pages={21707--21719},
  year={2025}
}

@inproceedings{duan2024vlmevalkit,
  title={Vlmevalkit: An open-source toolkit for evaluating large multi-modality models},
  author={Duan, Haodong and Yang, Junming and Qiao, Yuxuan and Fang, Xinyu and Chen, Lin and Liu, Yuan and Dong, Xiaoyi and Zang, Yuhang and Zhang, Pan and Wang, Jiaqi and others},
  booktitle={Proceedings of the 32nd ACM International Conference on Multimedia},
  pages={11198--11201},
  year={2024}
}

@inproceedings{vtw,
  title={Boosting multimodal large language models with visual tokens withdrawal for rapid inference},
  author={Lin, Zhihang and Lin, Mingbao and Lin, Luxi and Ji, Rongrong},
  booktitle={Proceedings of the AAAI Conference on Artificial Intelligence},
  pages={5334--5342},
  year={2025}
}

@inproceedings{geo3k,
  title={Inter-gps: Interpretable geometry problem solving with formal language and symbolic reasoning},
  author={Lu, Pan and Gong, Ran and Jiang, Shibiao and Qiu, Liang and Huang, Siyuan and Liang, Xiaodan and Zhu, Song-Chun},
  booktitle={Proceedings of the 59th Annual Meeting of the Association for Computational Linguistics and the 11th International Joint Conference on Natural Language Processing (Volume 1: Long Papers)},
  pages={6774--6786},
  year={2021}
}

@inproceedings{geoqa,
  title={Geoqa: A geometric question answering benchmark towards multimodal numerical reasoning},
  author={Chen, Jiaqi and Tang, Jianheng and Qin, Jinghui and Liang, Xiaodan and Liu, Lingbo and Xing, Eric and Lin, Liang},
  booktitle={Findings of the Association for Computational Linguistics: ACL-IJCNLP 2021},
  pages={513--523},
  year={2021}
}

@article{iconqa,
  title={Iconqa: A new benchmark for abstract diagram understanding and visual language reasoning},
  author={Lu, Pan and Qiu, Liang and Chen, Jiaqi and Xia, Tony and Zhao, Yizhou and Zhang, Wei and Yu, Zhou and Liang, Xiaodan and Zhu, Song-Chun},
  journal={arXiv preprint arXiv:2110.13214},
  year={2021}
}

@inproceedings{mathv360k,
  title={Math-llava: Bootstrapping mathematical reasoning for multimodal large language models},
  author={Shi, Wenhao and Hu, Zhiqiang and Bin, Yi and Liu, Junhua and Yang, Yang and Ng, See Kiong and Bing, Lidong and Lee, Roy Ka-Wei},
  booktitle={Findings of the Association for Computational Linguistics: EMNLP 2024},
  pages={4663--4680},
  year={2024}
}

@inproceedings{cmm_math,
  title={Cmm-math: A chinese multimodal math dataset to evaluate and enhance the mathematics reasoning of large multimodal models},
  author={Liu, Wentao and Pan, Qianjun and Zhang, Yi and Liu, Zhuo and Wu, Ji and Zhou, Jie and Zhou, Aimin and Chen, Qin and Jiang, Bo and He, Liang},
  booktitle={Proceedings of the 33rd ACM International Conference on Multimedia},
  pages={12585--12591},
  year={2025}
}

@inproceedings{olympiadbench,
  title={Olympiadbench: A challenging benchmark for promoting agi with olympiad-level bilingual multimodal scientific problems},
  author={He, Chaoqun and Luo, Renjie and Bai, Yuzhuo and Hu, Shengding and Thai, Zhen and Shen, Junhao and Hu, Jinyi and Han, Xu and Huang, Yujie and Zhang, Yuxiang and others},
  booktitle={Proceedings of the 62nd Annual Meeting of the Association for Computational Linguistics (Volume 1: Long Papers)},
  pages={3828--3850},
  year={2024}
}

@inproceedings{raven,
  title={Raven: A dataset for relational and analogical visual reasoning},
  author={Zhang, Chi and Gao, Feng and Jia, Baoxiong and Zhu, Yixin and Zhu, Song-Chun},
  booktitle={Proceedings of the IEEE/CVF conference on computer vision and pattern recognition},
  pages={5317--5327},
  year={2019}
}

@article{mm_iq,
  title={Mm-iq: Benchmarking human-like abstraction and reasoning in multimodal models},
  author={Cai, Huanqia and Yang, Yijun and Hu, Winston},
  journal={arXiv preprint arXiv:2502.00698},
  year={2025}
}

@article{easyarc,
  title={Easyarc: Evaluating vision language models on true visual reasoning},
  author={Unsal, Mert and Akkus, Aylin},
  journal={arXiv preprint arXiv:2506.11595},
  year={2025}
}

@inproceedings{MATSIR_acl26,
  title = {LLM Inductive Reasoning Through Multi-Agent Enhanced Monte Carlo Tree Search},
  author = {Xiang Li and others},
  booktitle = {ACL},
  year={2026}
}

@article{li2026can,
  title={Can large multimodal models understand agricultural scenes? benchmarking with agromind},
  author={Li, Qingmei and Zhang, Yang and Mai, Zurong and Chen, Yuhang and Huang, Henglian and Zhang, Jiarui and Zhang, Zhiwei and Wen, Yibin and Li, Weijia and Fu, Haohuan and others},
  journal={Advances in Neural Information Processing Systems},
  year={2026}
}

@article{du2026medfuse,
  title={MedFuse: a multi-source data fusion framework for diabetic retinopathy lesion segmentation},
  author={Du, Jiantong and Li, Yan and Li, Yawen and Liao, Liwen and Zhao, Zhihao and Ye, Guanhua},
  journal={Frontiers of Computer Science},
  year={2026}
}

@inproceedings{fu2026mergevla,
  title={Mergevla: Cross-skill model merging toward a generalist vision-language-action agent},
  author={Fu, Yuxia and Zhang, Zhizhen and Zhang, Yuqi and Wang, Zijian and Huang, Zi and Luo, Yadan},
  booktitle=CVPR,
  year={2026}
}

@article{wang2026commit,
  title={When to Commit? Towards Variable-Size Self-Contained Blocks for Discrete Diffusion Language Models},
  author={Wang, Danny and Qiu, Ruihong and Huang, Zi},
  journal={arXiv preprint arXiv:2604.23994},
  year={2026}
}

@inproceedings{yuan2025wiswheat,
  title={WisWheat: A Three-Tiered Vision-Language Dataset for Wheat Management},
  author={Yuan, Bowen and Song, Selena and Fernandez, Javier and Luo, Yadan and Baktashmotlagh, Mahsa and Wang, Zijian},
  booktitle={Proceedings of the 33rd ACM International Conference on Multimedia},
  year={2025}
}

@inproceedings{zhao2025continual,
  title={Continual text-to-video retrieval with frame fusion and task-aware routing},
  author={Zhao, Zecheng and Chen, Zhi and Huang, Zi and Sadiq, Shazia and Chen, Tong},
  booktitle={Proceedings of the 48th International ACM SIGIR Conference on Research and Development in Information Retrieval},
  year={2025}
}

@inproceedings{yigeng_acl25,
  title={Reflection on Knowledge Graph for Large Language Models Reasoning},
    author = {Yigeng Zhou and Wu Li and Yifan Lu and Jing Li and Fangming Liu and Meishan Zhang and Yequan Wang and Daojing He and Honghai LIU and Min Zhang},
  booktitle = {Findings of the 63rd Annual Meeting of the Association for Computational Linguistics (ACL)},
  year={2025}
}

@inproceedings{echoes_iclr26,
  title={Echoes as Anchors: Probabilistic Costs and Attention Refocusing in LLM Reasoning},
  author = {Zhuoyuan Hao and Zhuo Li and Wu Li and Fangming Liu and Min Zhang and Jing Li},
  booktitle = {The Fourteenth International Conference on Learning Representations (ICLR)},
  year={2026}
}

@article{sun2024massive,
  title={Massive activations in large language models},
  author={Sun, Mingjie and others},
  journal={arXiv},
  year={2024}
}

@inproceedings{wang2026language,
  title={Language-driven fine-grained retrieval},
  author={Wang, Shijie and Yu, Xin and Luo, Yadan and Wang, Zijian and Zhang, Pengfei and Huang, Zi},
  booktitle={Proceedings of the IEEE/CVF Conference on Computer Vision and Pattern Recognition},
  pages={2682--2692},
  year={2026}
}

\newpage

\appendix


\noindent This supplementary material provides additional descriptions of the main paper. It is organized into three parts: (1) implementation details covering training configuration, hyperparameter settings, and data resources used in our experiments; (2) supplementary experiments providing further quantitative validation of our method; (3) Limitations; and (4) qualitative analysis presenting model response visualizations.

\begin{itemize}
    \item \textbf{\Cref{sec:training_config}:} RL Training Configuration.
    \item \textbf{\Cref{sec:hyperparam_config}:} Hyperparameter Configuration.
    \item \textbf{\Cref{sec:sink_id}:} Visual Sink Token Identification.
    \item \textbf{\Cref{sec:overhead}:} Training Overhead.
    \item \textbf{\Cref{sec:data_resources}:} Data Resources.
    \item \textbf{\Cref{sec:benchmark_stats}:} Benchmark Statistics.
    \item \textbf{\Cref{sec:system_prompt}:} System Prompt.
    \item \textbf{\Cref{sec:supp_exp}:} Supplementary Experiments.
    \item \textbf{\Cref{sec:limitation}:} Limitations.
    \item \textbf{\Cref{sec:qualitative}:} Qualitative Analysis.
\end{itemize}

\section{Implementation Details}

\subsection{Training Configuration}
\label{sec:training_config}
We train our models using the verl framework~\cite{verl}, with source code provided in the supplementary materials. For RL training with proposed reward functions, we train Qwen2.5-VL-7B Instruct model using GRPO within the VERL framework. The vision encoder is frozen throughout RL training.
For rollout generation, we sample $n=8$ responses per sample, and temperature of 0.7 and top-$p$ of 0.95. The KL divergence applied with a coefficient of 0.02. Clipping is applied with a low clip ratio of 0.2 and a high clip ratio of 0.28.
During RL training, we set the learning rate to $2 \times 10^{-6}$ with a constant warmup scheduler and train using BF16 mixed precision. The training hyperparameters are summarized in Table~\ref{tab:rl_config}.

\begin{table}[ht]
\centering
\caption{Key hyperparameters for RL training.}
\label{tab:rl_config}
\begin{tabular}{ll}
\toprule
\textbf{Hyperparameter}              & \textbf{Value}            \\
\midrule
Learning rate                        & $2 \times 10^{-6}$        \\
LR scheduler                         & Constant warmup           \\
Precision                            & BF16                      \\
Train batch size                     & 512                       \\
Rollout responses per sample     &     8                         \\
Rollout temperature                  & 0.7                       \\
Rollout top-$p$                      & 0.95                      \\
Rollout top-$k$                      & 20                        \\
KL loss coefficient                  & 0.02                      \\
KL loss type                         & Low-variance KL           \\
Clip ratio low                       & 0.2                       \\
Clip ratio high                      & 0.28                      \\
Loss aggregation                     & Token-mean                \\
\bottomrule
\end{tabular}
\end{table}

\subsection{Hyperparameters Configuration}
\label{sec:hyperparam_config}
We summarize the key hyperparameters of our method in Table~\ref{tab:hyperparams}. Below we provide additional justification for the design choices behind selected parameters.

To prevent reward hacking, we introduce a length penalty controlled by $\beta$. Without this regularization, the model can exploit the attention reward by generating extensively long responses.
Longer output produces more decoding steps, which increases the cumulative potentials for high visual attention scores, thus inflating the reward without genuine improvements in visual grounding. The parameter $\beta$ is set to 0.01 to penalize excessive response length, discouraging this degenerate behavior while preserving the model's ability to produce sufficient reasoning chains.

$\lambda$ controls the slope and sensitivity of the attention reward function across decoding timesteps. 
Specifically, $\lambda$ controls the steepness of the reward transition, determining how sharply the reward signal responds to changes in visual attention proportion.

Computing the visual grounding reward $R_\text{vis}$ at each individual decoding step is statistically noisy and semantically unreasonable. As it is natural that certain tokens (\emph{e.g.}, punctuation, conjunctions, or reasoning connectives) are generated with low visual attention without indicating a failure of visual grounding.
To address this, we adopt a sliding window strategy to aggregate over a local temporal context.
We compute a smoothed attention signal by taking the mean VAP over a fixed size window of consecutive decoding steps. Practically, we set window size to be 20 tokens in all experiments. This design yields a more stable and semantically meaningful reward signal.

\begin{table}[ht]
\centering
\caption{Hyperparameters of the framework.}
\label{tab:hyperparams}
\begin{tabular}{cll}
\toprule
\textbf{Parameter} & \textbf{Value} \\
\midrule
$\eta$   & \texttt{[20]} \\
$\beta$  & \texttt{[0.01]} \\
$\lambda$ & \texttt{[5]} \\
$\gamma$ & \texttt{[0.5]} \\
$\omega$ & \texttt{[0.3]} \\
\bottomrule
\end{tabular}
\end{table}

\subsection{Visual Sink Token Identification}
\label{sec:sink_id}
Our protocol follow~\cite{find_visual_sink}. Concretely, (i) The anomalous-dimension set $\mathcal{D}^{*}$ is \emph{not} estimated from data:~\cite{find_visual_sink, sun2024massive} report that sink dimensions are inherited from the base LLM and remain unchanged after multimodal fine-tuning (\eg\ $\{458,2570\}$ for Qwen2).
(ii) A visual token $j$ is a sink token if its mean magnitude restricted to $\mathcal D^{\!*}$ is excessive relative to $\mathcal V$ at the final layer $l^{\!*}$. The final-layer choice follows~\cite{find_visual_sink}, which empirically shows massive-activation patterns are most concentrated in late layers.
This identification serves as a building block; LASER's contribution lies in the \textbf{sink-suppression reward}, which explicitly regulates attention away from sink tokens during training.

\subsection{Training Overhead}
\label{sec:overhead}
\textsc{LASER} introduces attention-based rewards that require access to the model's internal attention maps during rollout. It is incurred only at training time, and we quantify the training overhead in the verl framework~\cite{verl}. We perform one additional forward pass per rollout over the already generated trajectory. This auxiliary pass does not involve additional sampling, and it re-runs the fixed token sequence to obtain the attention.
Empirically, vanilla GRPO takes 1251\,s/step, while \textsc{LASER} takes 1395\,s/step.

\subsection{Data Resources}
\label{sec:data_resources}
We construct our GRPO training dataset by combining and filtering data from publicly available multimodal reasoning corpora: the MMR1-RL dataset~\cite{mmr1} and RL dataset from ReVisual-R1~\cite{revisual_r1}.
Both datasets were curated with explicit attention to quality, difficulty, and diversity 
of multimodal reasoning.
The summary of the constituent source datasets is provided in \cref{tab:data_resouce}.

\begin{table}[ht]
\centering
\caption{Summary of original source datasets used to construct the RL training data pool.}
\label{tab:data_resouce}
\begin{tabular}{lll}
\toprule
\textbf{Dataset} & \textbf{Domain} & \textbf{Sample Size} \\
\midrule
Geometry3K~\cite{geo3k}     & Plane geometry         & 1.5K \\
GeoQA~\cite{geoqa}             & Geometric numerical reasoning         & 1.2K \\
IconQA~\cite{iconqa}             & Diagram VQA                  & 8.3K \\
MathV360K~\cite{mathv360k}      & Multimodal math  & 4.3K   \\
CMM-Math~\cite{cmm_math}    & Multimodal math       & 3.6K \\
OlympiadBench~\cite{olympiadbench}  & Bilingual Olympiad science & 0.4K \\
Revisual-R1 Curated~\cite{revisual_r1} & Multimodal Reasoning & 11.5K \\
MMR1 Curated~\cite{mmr1} &  Multimodal Math & 8K  \\
Raven~\cite{raven}, MM-IQ~\cite{mm_iq}, EasyArc~\cite{easyarc} & Logical Reasoning & 7K \\
\bottomrule
\end{tabular}
\end{table}

\subsection{Benchmark Statistics}
\label{sec:benchmark_stats}
We evaluate our method across diverse benchmarks spanning math, general multimodal understanding, logical reasoning, and visual grounding. Below we provide a brief introduction to each benchmark.

\begin{itemize}

    \item \textbf{MathVista}~\cite{mathvista} is a consolidated mathematical reasoning benchmark within visual contexts. It consists of 6,141 examples derived from 28 existing multimodal datasets involving mathematics and 3 newly created datasets (IQTest, FunctionQA, and PaperQA).  The benchmark spans diverse task types including figure QA, geometry problem solving, math word problems, textbook QA, and visual QA. We evaluate on the standard \texttt{testmini} split of 1,000 examples.

    \item \textbf{MathVerse}~\cite{mathverse} is an all-around visual math benchmark designed for equitable and in-depth evaluation. It encompasses 2,612 visual math problems spanning plane geometry, solid geometry, and functions. Each problem is transformed by human annotators into six distinct versions with varying degrees of multimodal information content, yielding 15K test samples in total.  We follow standard practice and report accuracy on the \texttt{Vision-Only} split of the \texttt{testmini} set (788 problems), which maximally stresses visual grounding.

    \item \textbf{MathVision}~\cite{mathvision} (MATH-V) is a high-difficulty mathematical reasoning benchmark. It is a meticulously curated collection of 3,040 high-quality mathematical problems with visual contexts sourced from real math competitions, spanning 16 distinct mathematical disciplines and graded across 5 levels of difficulty. 

    \item \textbf{WeMath}~\cite{wemath} is the first benchmark specifically designed to probe problem-solving \emph{principles} beyond end-to-end performance. It meticulously collects 6.5K visual math problems and decomposes them into 10.9K step-level questions for evaluation, spanning 5 layers of knowledge granularity and 67 hierarchical knowledge concepts.  It introduces a four-dimensional diagnostic metric (IK / IG / CM / RM) to distinguish insufficient knowledge from inadequate generalization.

    \item \textbf{MMMU}~\cite{MMMU} is the Massive Multi-discipline Multimodal Understanding and Reasoning benchmark. It presents 11.5K college-level problems across six broad disciplines and 30 college subjects, featuring highly heterogeneous image types and requiring expert-level perception and reasoning rooted in deep subject knowledge.  We report accuracy on the standard validation split of 900 examples.

    \item \textbf{MMStar}~\cite{mmstar} is an elite vision-indispensable multimodal benchmark. It comprises 1,500 challenge samples meticulously selected by humans, benchmarking 6 core capabilities and 18 detailed axes.  Each sample is required to be visually dependent, minimally data-leaked, and to demand advanced multimodal capabilities, directly addressing shortcomings in prior evaluation suites.

    \item \textbf{LogicVista}~\cite{logicvista} is a comprehensive benchmark for evaluating general logical reasoning capabilities of MLLMs in visual contexts. It evaluates logical cognition abilities across 5 logical reasoning tasks encompassing 9 different capabilities, using a sample of 448 multiple-choice questions. Each question is annotated with the correct answer and human-written reasoning behind the selection, enabling both open-ended and multiple-choice evaluation.  The five task categories are deductive, inductive, spatial, numerical, and mechanical reasoning.

    \item \textbf{HallusionBench}~\cite{hallusionbench} is an advanced diagnostic suite targeting entangled language hallucination and visual illusion in LVLMs. The benchmark comprises 346 images paired with 1,129 questions, all meticulously crafted by human experts, and introduces a novel control-group structure to enable quantitative analysis of models' response tendencies, logical consistency, and failure modes. 

\end{itemize}

\subsection{System Prompt}
\label{sec:system_prompt}
We provide the structured prompt template to elicit step-by-step multimodal reasoning from the model. Specifically, the system prompt instructs the model to analyze the provided inputs before providing final answer, where the reasoning proces is enclosed withthin \texttt{<think>} \texttt{</think>} tags, and the final answer is then isolated within  \texttt{<answer>} \texttt{</answer>} tags.

\begin{promptbox}{Prompt Template for LASER Training}

\textbf{System:}

\smallskip
A conversation between User and Assistant. The User provides an image and asks a question. The Assistant first analyzes both the image and the question, then carefully thinks about the reasoning process step by step, and finally provides the User with an accurate answer. The Assistant must carefully check the correctness and validity of each reasoning step. If any errors or inconsistencies are found during the reasoning process, the Assistant reflects and corrects them logically. The reasoning process and answer are enclosed within \texttt{<think>} \texttt{</think>} and \texttt{<answer>} \texttt{</answer>} tags, respectively, i.e.,

\smallskip
\texttt{<think>} \textit{reasoning process here, with potential reflections and corrections} \texttt{</think>}\\
\texttt{<answer>} \textit{final answer here, with the key result enclosed in} \texttt{\textbackslash boxed\{\}} \texttt{</answer>}

\medskip
\textbf{User:}

\smallskip
\texttt{\{image\}} \texttt{\{question\}}



\end{promptbox}

\section{Supplementary Experiments}
\label{sec:supp_exp}

\subsection{Visual-Targeted Evaluation on MathVista}
\label{sec:mathvista_visual}
As reported in the main experiment table, \textsc{LASER}'s performance on the full MathVista benchmark falls short of several baseline models. We hypothesize that this gap is partially attributable to the composition of MathVista that a non-trival subset of its samples can be answered correctly using only textual inputs, without requiring any visual evidence.

To investigate this, we conduct a targeted evaluation, in which we first identify samples answerable from text alone by evaluating all models under a \textit{text-only} setting. Samples for which the models answer correctly without visual input are excluded, resulting in a Visual-Targeted subset that requires image understanding to solve. We then re-evaluate all models on this filtered subset.

As shown in \cref{tab:mathvista_visual}, while all models exhibit a performance drop on the Visual-Targeted subset, \textsc{LASER} outperforms baseline models, achieving 65.0 accuracy and the smallest performance dropping rate. The results indicate that \textsc{LASER} enhances the model's ability to attend to and reason over visual evidence, steering the model toward image-based reasoning rather than answering driven by textual context alone.

\begin{table}[t]
\centering
\caption{Performance comparison on Visual-Targeted MathVista 
(text-solvable samples removed). Drop Rate denotes the accuracy decrease 
from the full to the visual-targeted subset.}
\label{tab:mathvista_visual}
\resizebox{\columnwidth}{!}{%
\begin{tabular}{l cc cc cc}
\toprule
\multirow{2}{*}{\textbf{Dataset}} 
    & \multicolumn{2}{c}{\textbf{VisionR1-7B}} 
    & \multicolumn{2}{c}{\textbf{OpenVLThinker-7B}} 
    & \multicolumn{2}{c}{\textbf{LASER}} \\
\cmidrule(lr){2-3} \cmidrule(lr){4-5} \cmidrule(lr){6-7}
    & Acc. & Drop$\downarrow$
    & Acc. & Drop$\downarrow$
    & Acc. & Drop$\downarrow$ \\
\midrule
Visual-Targeted MathVista         & 61.5 & 16.3\%  & 60.2 & 16.7\%  & \textbf{65.0} & \textbf{14.9\%}  \\
\bottomrule
\end{tabular}%
}
\end{table}

\subsection{Sensitivity Analysis on $\lambda$}
\label{sec:sensitivity}
To assess robustness, we evaluate $\lambda \in \{1, 2.5, 5, 7.5, 10\}$ while all other hyperparameters are fixed, and the average accuracy across the benchmarks is shown in \cref{tab:lambda}. Performance is stable across the range.

\begin{table}[ht]
\centering
\caption{Sensitivity to $\lambda$: average accuracy across benchmarks.}
\label{tab:lambda}
\begin{tabular}{cccccc}
\toprule
$\lambda$ & 1 & 2.5 & 5 & 7.5 & 10 \\
\midrule
Accuracy & 52.4 & 53.3 & \textbf{54.6} & 53.9 & 53.2 \\
\bottomrule
\end{tabular}
\end{table}

\subsection{Early-Weighted Variant of $R_{\text{vis}}$}
\label{sec:early_weight}
\textbf{Finding~1} of the main paper identifies attention \emph{decay along the generation horizon} as the core pathology: early-stage grounding establishes the semantic scaffold for subsequent reasoning, and errors there propagate autoregressively. We also test whether $R_{\text{vis}}$ should therefore weight early decoding steps more heavily. We evaluate an explicit early-weighted variant that re-weights the per-step reward with an exponentially decaying schedule. As shown in \cref{tab:early_weight}, the early-weighted variant performs comparably to the uniform default (52.2 vs.\ 52.3 on average).

\begin{table}[ht]
\centering
\caption{Early-weighted vs.\ uniform (peak-normalized) $R_{\text{vis}}$. Both variants are trained identically and evaluated on the same VLMEvalKit pipeline, differing only in the per-step weighting $w_t$. The two are comparable, so we adopt the simpler uniform form as the default.}
\label{tab:early_weight}
\resizebox{\columnwidth}{!}{%
\begin{tabular}{lccccccc}
\toprule
\textbf{Variant} & MathVista & MathVerse & MathVision & MMStar & LogicVista & Hallu.B & \textbf{Avg.} \\
\midrule
Early-weighted ($w_t{=}e^{-t/T}$) & \textbf{70.8} & 44.9 & 26.4 & \textbf{65.3} & 50.0 & 55.8 & 52.2 \\
Uniform (default)                 & 69.7 & \textbf{45.6} & \textbf{27.4} & 64.3 & \textbf{50.3} & \textbf{56.6} & \textbf{52.3} \\
\bottomrule
\end{tabular}%
}
\end{table}

\section{Limitations}
\label{sec:limitation}
Despite effectiveness of \textsc{LASER}, our method has several limitations that point to directions for future work. First, due to computational constraints, we only conduct experiments at the 7B parameters scale using Qwen2.5-VL-7B-Instruct as the baseline model. While this scale is sufficient to validate our core contributions, the exploration for scaling to larger LVLMs remains an open question. Scaling \textsc{LASER} to larger LVLMs and more model families is an important direction for future work. Second, visual sink token identification relies on architecture-specific attention patterns, potentially requiring recalibration when applied to other model families with different model structures. Developing architecture-agnostic criteria for sink detection presents a direction for future work.

\section{Qualitative Analysis}
\label{sec:qualitative}
\begin{figure*}[!t]
\centering
    \includegraphics[width=\textwidth]{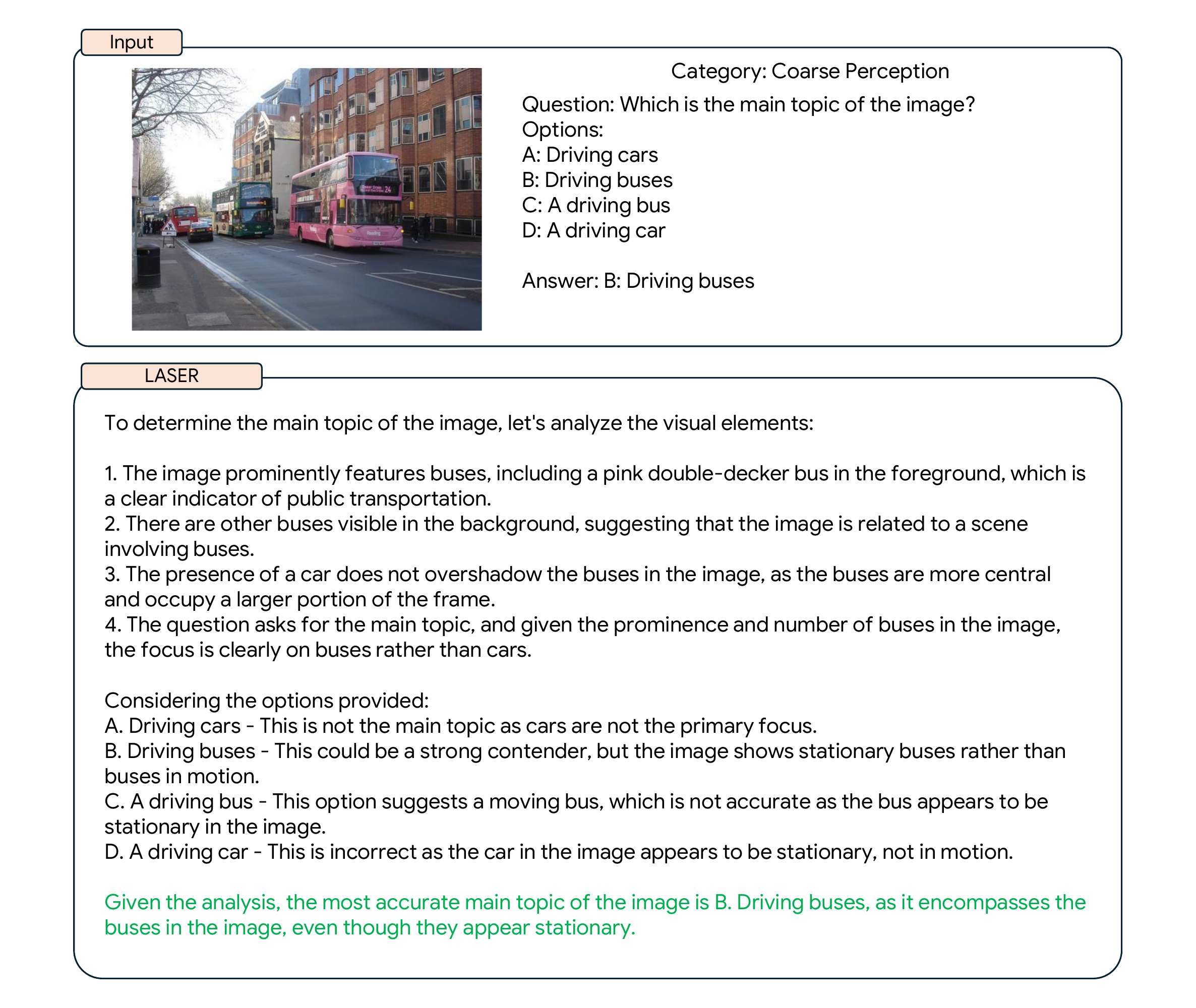}
    \caption{Qualitative results of \textsc{LASER}.}
\label{fig:example1}
\end{figure*}

\begin{figure*}[!t]
\centering
    \includegraphics[width=\textwidth]{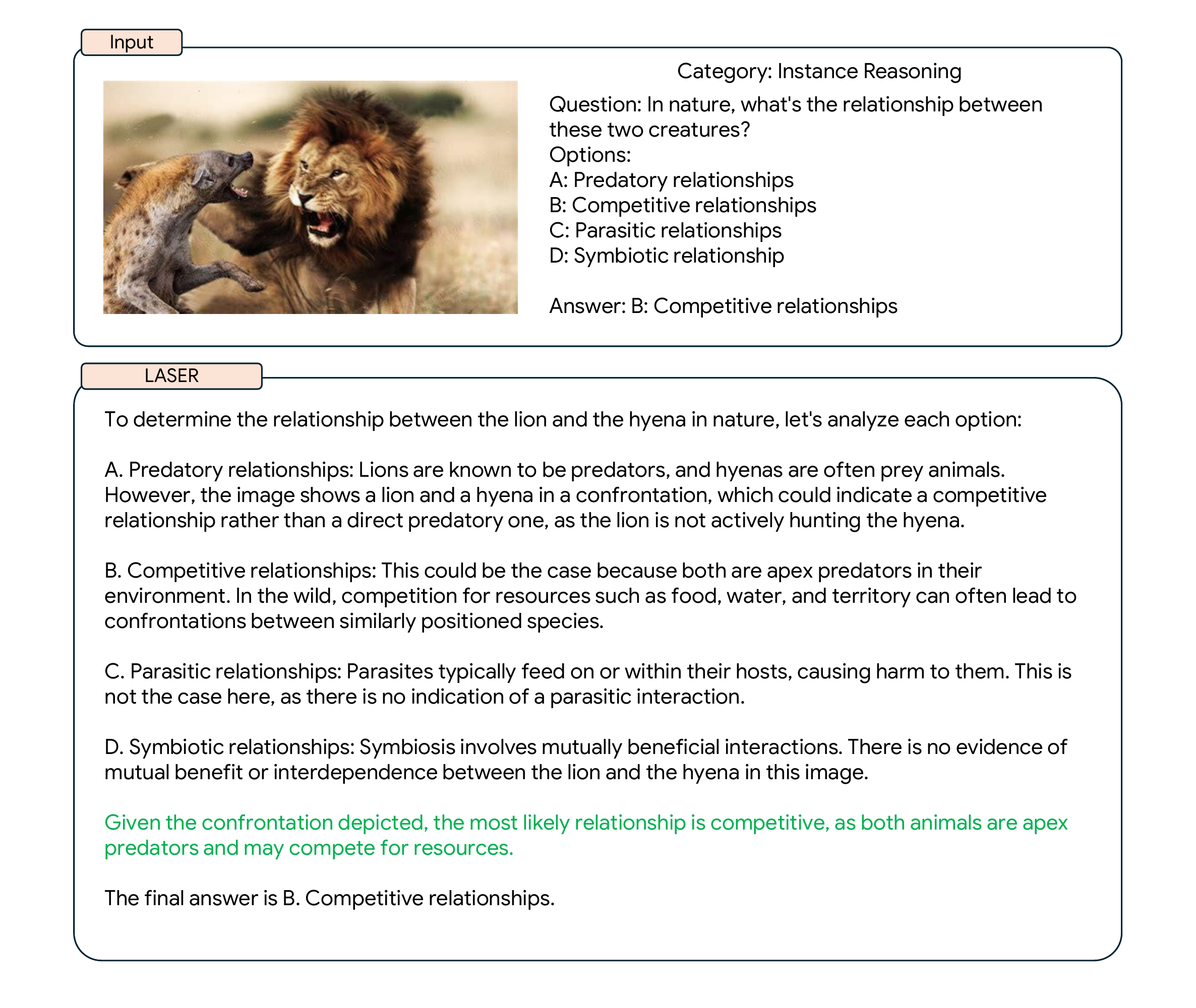}
    \caption{Qualitative results of \textsc{LASER}.}
\label{fig:example2}
\end{figure*}

\begin{figure*}[!t]
\centering
    \includegraphics[width=\textwidth]{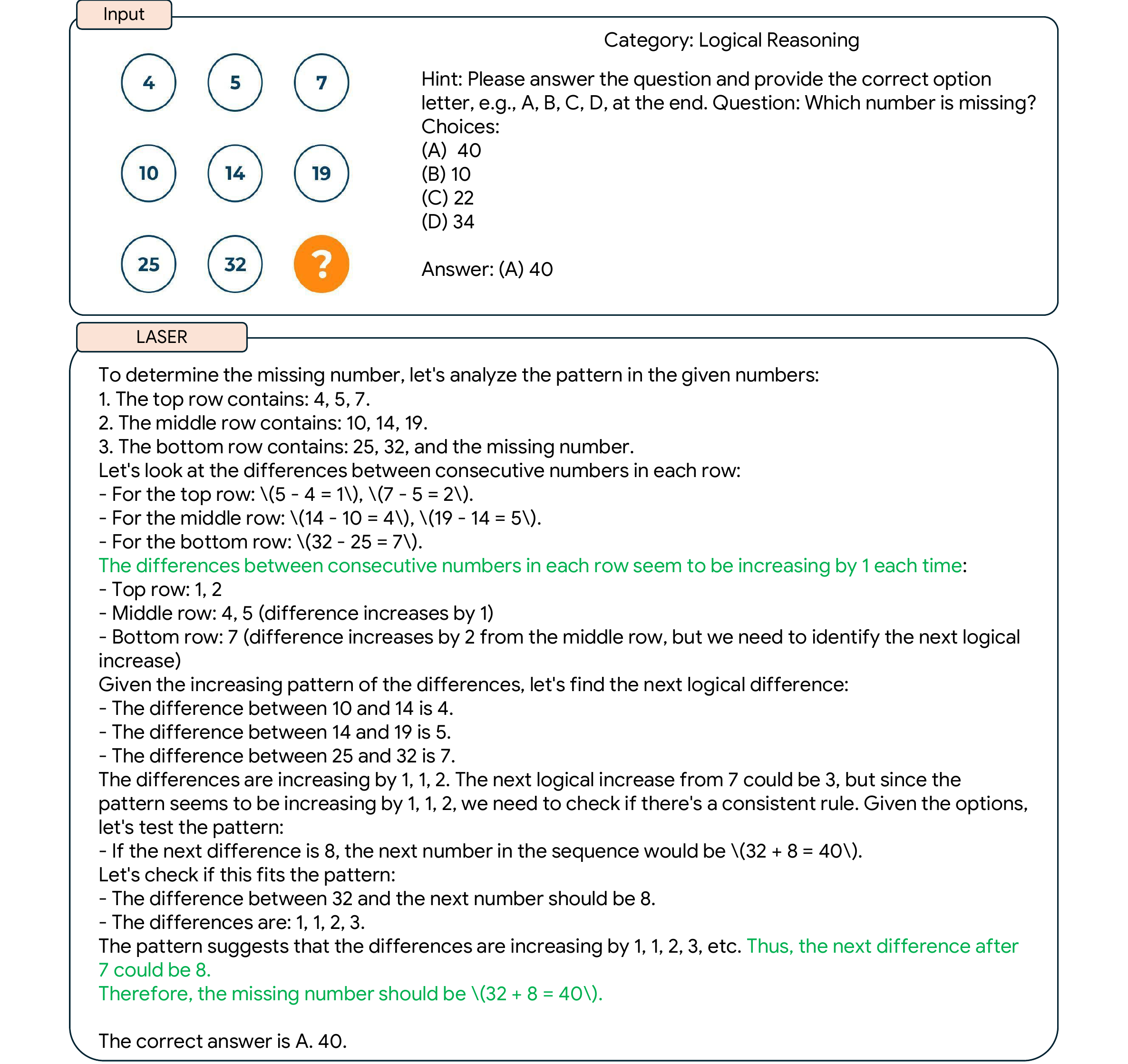}
    \caption{Qualitative results of \textsc{LASER}.}
\label{fig:example3}
\end{figure*}

\begin{figure*}[!t]
\centering
    \includegraphics[width=\textwidth]{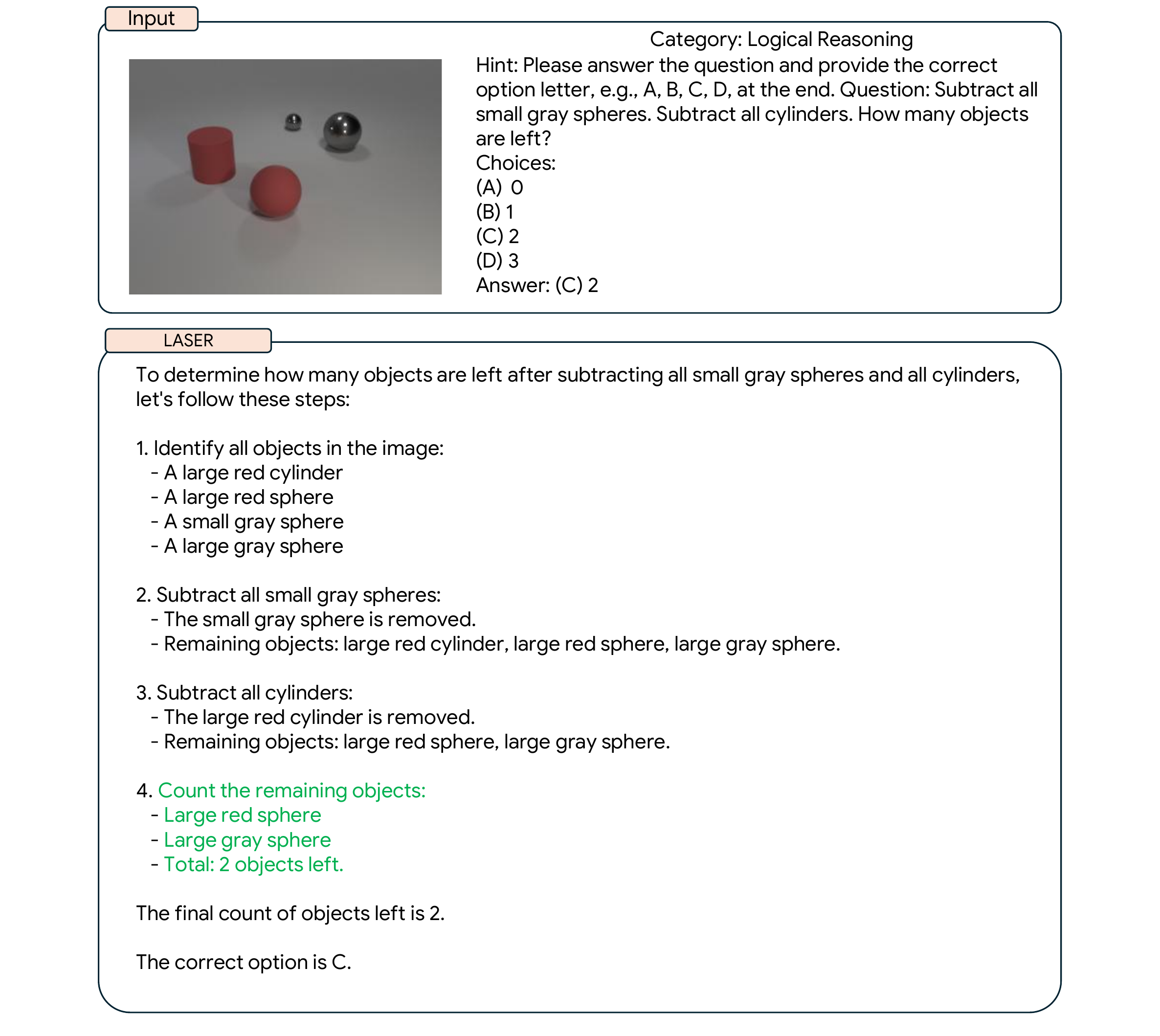}
    \caption{Qualitative results of \textsc{LASER}.}
\label{fig:example4}
\end{figure*}

To further illustrate the effectiveness of \textsc{LASER}, we present case studies showing the model's reasoning process on representative visual reasoning tasks. \cref{fig:example1} to \cref{fig:example4} display samples from the evaluation benchmarks, where we visualize the full chain-of-thought generated within the \texttt{<think>} \texttt{</think>} 
tags alongside the corresponding input image and final answer.

A consistent pattern emerges across all cases. \textsc{LASER} grounds its reasoning in the visual content from the outset, explicitly depicting the image content and identifying key visual elements before proceeding to logical inference.
This visual engagement is a direct reflection of \textsc{LASER}'s training objective: by mitigating attention drift toward visual sink tokens via 
$R_{\text{supp}}$ and mainting visual attention via $R_{\text{vis}}$, the model 
learns to reason \textit{with} the image rather than bypass it.

\clearpage
\end{document}